\documentclass[smallextended]{svjour3}
\smartqed  
\usepackage{graphicx} 
\usepackage{subfigure}
\usepackage{array}
\usepackage{amsmath,amssymb,amsfonts}
\usepackage{slashbox,multirow}
\usepackage{pdflscape}
\usepackage{afterpage}
\usepackage{footnote}

\usepackage{longtable}

\usepackage{booktabs}

\usepackage{algorithm}
\usepackage{algorithmic}

\usepackage{natbib}

\usepackage{url}
\usepackage{hyperref}

\newcommand {\bs}[1] {\boldsymbol{#1}}
\newcommand{\bbR}{\mathbb{R}}

\newcommand {\mat}       {\begin{bmatrix}}
\newcommand {\rix}          {\end{bmatrix}}
\newcommand {\matt}      {\begin{pmatrix}}
\newcommand {\rixx}          {\end{pmatrix}}

\begin{document}

\title{Multivariate Regression with Grossly Corrupted Observations: A Robust Approach and its Applications
}

\titlerunning{Multivariate Regression with Grossly Corrupted Observations}        

\author{Xiaowei~Zhang \and 
        Chi~Xu \and 
        Yu~Zhang \and 
        Tingshao~Zhu \and 
        Li~Cheng
        }

\institute{Xiaowei~Zhang \and Chi~Xu \and Yu~Zhang \at
           Bioinformatics Institute, A*STAR, Singapore. \\
           \email{\{zhangxw, xuchi, zhangyu\}@bii.a-star.edu.sg} \and
           Tingshao~Zhu \at
           Institute of Psychology, Chinese Academy of Sciences, China.\\
           \email{tszhu@psych.ac.cn} \and
           Li~Cheng \at
           Bioinformatics Institute, A*STAR, Singapore and School of Computing, National   University of Singapore, Singapore. \\
           \email{chengli@bii.a-star.edu.sg}
           }

\date{}

\maketitle

\begin{abstract}
This paper studies the problem of multivariate linear regression where a portion of the observations is grossly corrupted or is missing, and the magnitudes and locations of such occurrences are unknown in priori. To deal with this problem, we propose a new approach by explicitly consider the error source as well as its sparseness nature. An interesting property of our approach lies in its ability of allowing individual regression output elements or tasks to possess their unique noise levels. Moreover, despite working with a non-smooth optimization problem, our approach still guarantees to converge to its optimal solution.
Experiments on synthetic data demonstrate the competitiveness of our approach compared with existing multivariate regression models.
In addition, empirically our approach has been validated with very promising results on two exemplar real-world applications: The first concerns the prediction of \textit{Big-Five} personality based on user behaviors at social network sites (SNSs), while the second is 3D human hand pose estimation from depth images.
The implementation of our approach and comparison methods as well as the involved datasets are made publicly available in support of the open-source and reproducible research initiatives.
\end{abstract}

\section{Introduction}\label{sec:intro}
The multivariate linear model, also known as general linear model, plays an important role in multivariate analysis~\citep{Anderson03}. It can be written as
\begin{equation}\label{eq:MLR}
Y=XW^{*}+Z.
\end{equation}
Here $Y\in\bbR^{n\times p}$ is a matrix with a set of multivariate observations, $X\in\bbR^{n\times d}$ is referred to as a design matrix with each row being a sample, $W^{*}\in\bbR^{d\times p}$ is a regression coefficient matrix which needs to be estimated, and $Z\in\bbR^{n\times p}$ is a matrix containing observation noises. One of the central problems in statistics is to precisely estimate $W^{*}$ , the coefficient regression matrix from design matrix $X$ and noisy observations $Y$. It is typical to assume that the noise in $Y$ has bounded energy, and can be well-absorbed into the noise matrix $Z$. This is usually modelled to follow certain Gaussian-type distributions. It thus gives rise to the following regularized loss minimization framework in statistics, where $W^{*}$ is estimated by
\begin{equation}\label{eq:MLR_frame}
\hat{W}=\text{arg}\min_{W}\big\{\ell(X,Y;W)+\lambda r_W(W)\big\}.
\end{equation}
Here $\ell$ denotes a loss function, $\lambda>0$ refers to a tuning parameter, and $r_W(W)$ corresponds to a regularization term.
Moreover for $\ell$, the least square loss is usually the most popular choice, which has been shown to achieve the optimal rates of convergence under certain conditions on $X$ and $Z$~\citep{Lounici2011,Rohde2011}. It has also been applied in many applications including e.g. multi-task learning~\citep{ArgyriouEP08,Caruana97}.

Nonetheless, there exist real-life situations where certain entries in observation $Y$ are corrupted by considerably larger errors than those of the ``normal" ones that can be incorporated in the noise model considered above. Consider for example the following scenario: a few of data entries could be severely contaminated due to careless or even malicious user annotations, while these errors are unfortunately difficult to be identified in practice. This type of sparse yet large-magnitude noises might seriously damage the performance of the above-mentioned estimator $\hat{W}$.

Here we consider to tackle this problem of grossly corrupted observations \emph{explicitly} by considering a sparse matrix $G^{*}\in\bbR^{n\times p}$ with the locations of nonzero entries being unknown and with their magnitudes being possibly very large. This gives rise to a multivariate linear model as
\begin{equation}\label{eq:LinearMod_Gross}
Y=XW^{*}+Z+G^{*}.
\end{equation}
It thus enables us to restore those examples with gross errors instead of merely throwing them away as outliers. Note the same model is also capable of dealing with the missing data problem, i.e. situations in which a subset of observations in $Y$ are missing. More concretely, the missing observations can be imputed with zeros, then model \eqref{eq:LinearMod_Gross} is applied to this modified data. As a result, for each missing entry $Y_{ij}$ in $Y$ whose corresponding entry $G^{*}_{ij}$ in $G^{*}$ is nonzero, its negative $-G^{*}_{ij}$ forms the predicted correction.

This naturally leads to the following optimization framework of estimating $\left( W^{*}, G^{*} \right)$
\begin{equation}\label{eq:MLR_frameGross}
\min_{W, G} \big\{\overline{\ell}(X,Y;W,G)+\lambda r_W(W) +\rho r_G(G)\big\}
\end{equation}
with $\overline{\ell}$ being a loss function, $\rho>0$ as a trade-off parameter, and $r_G(G)$ as a regularization term.
Further, rather than the usual least square loss, the $\ell_{2,1}$-norm, defined as the sum of 2-norm of all columns, is considered here as the loss function due to its ability of dealing with different noise levels in regression tasks. Additionally, we employ a group sparsity inducing norm for $r_W(W)$ to enforce group-structured sparsity, and use $\ell_1$-norm for $r_G(G)$ to impose element-wise sparsity constraint for detecting possible gross corruptions.

This paper contains the following major contributions:
(1) A new approach is proposed in the context of multivariate linear regression to explicitly model and recover the missing or grossly corrupted observations, while the adoption of $\ell_{2,1}$-norm loss function also facilitates the ability of modeling the noise levels of individual outcome variables;
(2) The induced non-smooth optimization problem is addressed by our proposed multi-block proximal alternating direction method of multipliers (ADMM) solver, which is shown in this paper to be efficient and globally convergent;
(3) To demonstrate the general applicability of our approach, two interesting and distinct real-world applications have been examined: The first application involves the investigation of Big-Five personality from user online behaviors when interacting with social network sites (SNSs), an emerging problem from computational psychology. The second application concerns the challenging computer-vision problem of depth-image based human hand pose estimation. These two problems exemplify the broad spectrum of applications where our approach could be applicable. Empirical evaluation is carried out with synthetic and real datastets for both applications, where our approach is shown to compare favorably with existing multivariate regression models as well as application-specific state-of-the-art methods.
(4) Last but not least, to support the open-source and reproducible practice, our implementations and related datasets are also made publicly available~\footnote{Implementations of our approach as well as comparison methods, related datasets, and detailed information can be found at our dedicated project webpage~\url{http://web.bii.a-star.edu.sg/~zhangxw/cmrg}.}.

We note in the passing that part of this paper was presented in~\citep{ZhangEtAl:15}. Meanwhile, this paper presents substantial amount of new contents comparing to~\citep{ZhangEtAl:15}: From algorithmic aspect, our approach is carefully presented with more details and in a self-complete manner, with convergence analysis and proof, as well as time complexity analysis; From empirical application point of view, our approach is systematically examined on a series of simulated data. Practically, our approach has also been additionally validated on the interesting yet challenging application of depth-image based human hand pose estimation, where very competitive performance is obtained on both synthetic and real datasets. Besides, our code is also made publicly available in support of the open-source research practice; From presentation side, the paper is significant re-written and re-organized to accommodate the new materials, including e.g. review of hand pose estimation related literature; Overall the work presented in this paper is a lot more self-complete, and is better connected with real-world problems.

\subsection{Related Work}
In the line of work in machine learning and statistics, there have been various methods~\citep{Bhatia15,LiCS2012,NguyenTran13,WrightMa10,XuCM13,XuCS12,XuAISTATS12} proposed for linear regression with gross errors as in \eqref{eq:LinearMod_Gross}, among which \citep{NguyenTran13} and \citep{XuAISTATS12} examine the univariate and multivariate outputs, respectively, of the optimization problem \eqref{eq:MLR_frameGross}, where $\overline{\ell}$ is the standard least square loss. 
Nevertheless, as pointed out in \citep{Liu2014}, the least square loss has two drawbacks:
First, all regression tasks (each column in $Y$ or $W$ of~\eqref{eq:MLR}, which corresponds to an element of the multivariate regression output, can be regarded as a task) share the same regularization trade-off $\lambda$.
As a result, the varying noise levels contained in different regression tasks are unfortunately ignored;
Second, to improve the finite-sample performance, it is often important to choose an optimal $\lambda$ that depends on the estimation of unknown variance of $Z$.
Aiming at address these two issues, a calibrated multivariate regression (CMR) method has been proposed in \citep{Liu2014}, where the $\ell_{2,1}$-norm is employed as a novel loss function. It enjoys a number of desirable properties including being tuning insensitive and being capable of calibrating tasks according to their individual noise levels. Theoretical and empirical evidence~\citep{Liu2014} has demonstrated the ability of CMR to deliver an improved finite sample performance. This inspires us to adopt in our approach the $\ell_{2,1}$-norm as our loss function.
It is worth pointing out that our induced optimization problem and subsequently our proposed solver bear clearly differences from that of~\citep{Liu2014}.

\paragraph{Related Work in Personality Prediction}
So far, there are relatively few research efforts attempting toward personality prediction from SNSs behaviors~\citep{MaJiaZha:APS11}. One prominent theory in the study of personality is the Big-Five theory~\citep{Fun:ARP01,MatDeaWhi:book06}, which describes personality trait by five disparate dimensions, which are\emph{Conscientiousness},  \emph{Agreeableness}, \emph{Extraversion},
\emph{Openness}, and \emph{Neuroticism}. Traditionally, the most common way to predict an individual's personality is applying the \emph{self-report inventory}~\citep{psytestintro06}, which relies on the subjects to fill up questionnaires by themselves on their own behaviors, which are then summarized into a quantitative five-dimensional personality descriptor. However, such method has two disadvantages. First, it is practically infeasible to conduct self-report inventory in large-scale. Second, maliciously wrong answers might be supplied, or sometimes idealized answers or wishes are provided instead of the real ones, which nevertheless reduce the annotation credibility of the their behaviors. The above mentioned issues would collectively lead to highly deviated or sometimes missing personality descriptors. 

It has been widely accepted in psychology that an individual's personality can be manifested by behaviors. In recent years, these social networks including Facebook, Twitter, RenRen, and Weibo, have drastically changed the way people live and communicate. There have been evidences~\citep{LanLou:CHB06} suggesting that social network behaviors are significantly correlated with the real-world behaviors. On one hand, the fast growing number of SNSs users provides large amount of data for social research~\citep{Jie:STE11,Reynol:CE11}. On the other hand, there still lacks a proper model which can fully exploit the data to perform personality prediction. One research effort along this direction is probably that of Gosling et al.~\citep{GosEtAl:CBSN11}, which proposes a mapping between personality and SNSs behaviors. Specifically, they design 11 features, including friends count and weekly usage, based on \emph{self-reported} Facebook usage and observable profile information, and investigate the correlation between personality and these features. Meanwhile, these features entirely rely on statistical descriptions rather than ones that explicitly revealing user behaviors. Moreover, their data collection requires considerable manual efforts since the procedure is based on self-reported facebook utilization and online profile information, making it non-realistic for practical purpose.

\paragraph{Related Work in Hand Pose Estimation from Single Depth Images}
3D hand pose estimation~\citep{ErolEtAl:cviu07,GorFlePar:pami11} refers to the problem of estimating the finger-level human hand joint locations in 3D.
Vision-based hand interpretation has played important roles in diverse applications including humanoid animation~\citep{SueKauPai:siggraph08,WanPop:siggraph09}, robotic control~\citep{GusEtAl:BioCyber12}, and human-computer interaction~\citep{HacMccBro:vr11}, among others.
In its core lies this interesting yet challenging problem of 3D hand pose estimation, owing mostly to the complex and dexterous nature of hand articulations~\citep{GusEtAl:BioCyber12}.
Facilitated by the emerging commodity-level depth cameras, recent efforts such as~\citep{KesEtAl:eccv12,TanEtAl:cvpr14,XuEtAl:IJCV15} have led to noticeable progress in the field.
A binary latent tree model is used in~\citep{TanEtAl:cvpr14} to guide the searching process of 3D locations of hand joints, while
\citep{OikLouArg:cvpr14} adopts an evolutionary optimization method to capture hand and object interactions.
A dedicated random forest variant for hand pose estimation problem is proposed in~\citep{XuEtAl:IJCV15} with state-of-the-art empirical performance as well as nice theoretically consistency guarantees.
%
As an emerging research topic, the NYU Hand pose dataset~\citep{tompson14tog} is becoming the de facto benchmark for new methods to assess their performance on 3D hand pose estimation, which is also considered during the empirical evaluation section of our paper.

\subsection{Notations and Definitions}
Several notations and definitions are provided below. Given any scalar $\alpha$, we define $(\alpha)_{+}:=\max\{\alpha,0\}$, that is $(\alpha)_{+}=\alpha$ if $\alpha >0$ and 0 otherwise. Given a $\bs{x}=(x_1,\cdots,x_d)\in\bbR^d$ and $1\leq p < \infty$, we denote by $\|\bs{x}\|_p:=(\sum_{i=1}^{d}|x_i|^p)^{\frac{1}{p}}$ the $\ell_p$-norm and $\|\bs{x}\|_\infty:=\max_{i=1}^{d}|x_i|^p$ the $\ell_\infty$-norm. A group $g$ is a subset of $\{1,\cdots,d\}$ with cardinality $|g|$, while $\mathcal{G}$ denotes a set of groups, where each element $g$ corresponds to a group that potentially overlaps with others. Overall we have $\cup_{g\in\mathcal{G}}g=\{1,\cdots,d\}$. $\bs{x}_g$ denotes the subset of entries of $\bs{x}\in\bbR^d$ with indices in $g$. In a similar way, given a matrix $A\in \bbR^{n\times d}$, $A_{g*}$ and $A_{*g}$ refer to the rows and columns of $A$ indexed by $g$, respectively. An identity matrix $I$ is used with its size being self-explained from the context. $\mathcal{S}^{p}_{++}$ denotes the set of symmetric positive definite matrices of size $p$-by-$p$. In what follows we define three norms in a row, which are the Frobenius, spectral, and $\ell_{\infty}$-norms: $\|A\|_F:=\sqrt{\sum_{ij}A_{ij}^2}$, $\|A\|_2:= \max_{1\leq i \leq r}\sigma_{i}(A)$, and $\|A\|_{\infty}:=\max_{ij}|A_{ij}|$. $r$ denotes the matrix rank in our context, and $\sigma_i(A)$ denotes the $i$-th largest singular value of matrix $A$. We also define the matrix $\ell_{1,2}$-norm and $\ell_{2,1}$-norm as $\|A\|_{1,2}:=\sum_{i=1}^n\|A_{i*}\|_2$ and $\|A\|_{2,1}:=\sum_{j=1}^d\|A_{*j}\|_2$, respectively.
At last, the group lasso penalty associated with a group set $\mathcal{G}$ is defined as $\mathcal{R}_{\mathcal{G}}(A):=\sum_{g\in\mathcal{G}}\|A_{g*}\|_F$.

\subsection{Organization}
The rest of this paper is organized as follows. In Section \ref{sec:model}, we propose our robust multivariate regression model and compare its finite-sample performance with other multivariate regression models. In Section \ref{sec:Algo}, we describe the derivation of algorithm CMRG, which solves the induced optimization problem of our model via proximal ADMM, and provide its convergence analysis and complexity analysis. In Section \ref{sec:exp}, we evaluate the effectiveness of the proposed method on both synthetic and real data, and apply our method to predict personality from user behaviours at SNSs as well as estimate hand pose from depth images. Finally, conclusions are drawn in Section \ref{sec:concl}.

\section{The Proposed Model}\label{sec:model}
In our context, $r_W(W)=\mathcal{R}_{\mathcal{G}}(W)$ is denoted as a group sparsity inducing norm. The stochastic noise $Z$ considered in model \eqref{eq:MLR} is assumed to follow the following law $Z_{i*}\overset{\text{i.i.d}}{\sim} N(0,\Sigma)$ with a covariance matrix $\Sigma\in \mathcal{S}^{p}_{++}$.

To start with, we consdier a least square loss and its usage in the ordinary multivariate regression or OMR model, which gives rise to a convex optimization problem as follows:
\begin{equation}\label{eq:ReqMLR}
\hat{W}=\text{arg}\min\limits_{W} \left\{ \|Y-XW\|_F^2 + \lambda \mathcal{R}_{\mathcal{G}}(W) \right\}.
\end{equation}
Theoretically, it has been shown in \citep{Lounici2011} that, under the assumption that
$\Sigma=\text{diag}(\sigma_1^2,\cdots,\sigma_p^2)$ and suitable conditions on $X$, let $\sigma_{\max}=\max_{1\leq k\leq p} \sigma_k$, if we choose $\lambda=2c_1\sigma_{\max}(\sqrt{n\ln d} +\sqrt{np})$ for some $c_1>1$,
then with the following rate of convergence
\begin{align}\label{eq:Opt_con}
\frac{1}{\sqrt{p}}\|\hat{W} - W^{*} \|_F = O_p \left(\sqrt{\frac{s\ln d}{np}} + \sqrt{\frac{s}{n}} \right),
\end{align}
we obtain the estimator $\hat{W}$ of \eqref{eq:ReqMLR}. Here $s$ denotes the number of non-zeros rows in $W^*$ and $O_p$ means stochastic upper bond. However, as pointed out in \citep{Liu2014}, the empirical loss function of \eqref{eq:ReqMLR} has two potential drawbacks: (1) All the tasks are regularized by the same parameter $\lambda$, which introduces unnecessary estimation bias to some of the tasks with less noise in order to compensate the other tasks having larger noise. (2) It is highly non-trivial to decide a proper tuning parameter for a satisfactory result. 

As a remedy, Liu et al.~\citep{Liu2014} advocate a calibrated multivariate regression (CMR) model that uses $\ell_{2,1}$-norm as loss function, where where the regularization is dedicated to each regression task $k$. In other words, it is calibrated toward the individual noise level $\sigma_k$. Mathematically, CMR considers the following optimization problem:
\begin{equation}\label{eq:CLR}
\hat{W}=\text{arg}\min\limits_{W} \left\{\|Y-XW\|_{2,1} + \lambda \mathcal{R}_{\mathcal{G}}(W) \right\}.
\end{equation}
An re-interpretation of this CMR model in \eqref{eq:CLR} is from the following weighted regularized least squares problem:
\[
\min\limits_{W}\sum_{k=1}^{p}\frac{1}{\sigma_k}\|Y_{*k} - X W_{*k}\|_2^2 + \lambda\mathcal{R}_{\mathcal{G}}(W),
\]
where $1 / \sigma_k$ is the weight assigned to calibrate the $k$-th task. When there is no prior knowledge on $\sigma_k$, we estimate it by letting $\sigma_k=\|Y_{*k} - X W_{*k}\|_2$, which can be considered as the error in the $k$-th task. Theoretically, \citep{Liu2014} has proven that the CMR model \eqref{eq:CLR} achieves better finite-sample performance than the ordinary multivariate linear regression model \eqref{eq:ReqMLR} in the sense that estimator $\hat{W}$ achieves the optimal rates of convergence in \eqref{eq:Opt_con} if we choose $\lambda = c_2 (\sqrt{\ln d} + \sqrt{p})$, which is independent of $\sigma_k$. Therefore, the tuning parameter $\lambda$ in the OMR model depends on the noise level $\sigma_k$'s (through $\sigma_{\mathrm{max}}$), while the tuning parameter $\lambda$ in the CMR model is insensitive to the noise level $\sigma_k$'s.

Unfortunately, neither the OMR model of \eqref{eq:ReqMLR} nor the CMR model of~\eqref{eq:CLR} addresses the gross error issue. A natural idea toward addressing this problem is to explicitly model the gross errors in $Y$ as stated in \eqref{eq:LinearMod_Gross}. Then corrected observations can be obtained by simply removing gross errors from $Y$, which are further used to estimate coefficient regression matrix $W^{*}$. More specifically, we consider joint-forcing the benefits of both the CMR model of~\eqref{eq:CLR} and the model of~\eqref{eq:LinearMod_Gross} by investigating the optimization objective as
\begin{align}\label{eq:CMR_Gross}
(\hat{W},\hat{G}) = \text{arg}\min\limits_{W,G} \left\{ \|Y-XW-G\|_{2,1}+\lambda \mathcal{R}_{\mathcal{G}}(W)+\rho\|G\|_1 \right\}.
\end{align}
Here we have two regularization parameters, $\lambda>0$ and $\rho>0$. In particular, when letting $\rho = \infty$, we get $\hat{G} = 0$, and optimization problems~\eqref{eq:CLR} and~\eqref{eq:CMR_Gross} give the same solutions. In other words, when there is no gross error, our method reduces to the original CMR. In this regard, our method can be considered as an extension of CMR to deal with gross errors.

A related work is \citep{XuAISTATS12} that also looks at grossly corrupted observations but in the context of multi task regression. Its associated optimization problem could be reformulated as:
\begin{align}\label{eq:OMR_Gross}
(\hat{W},\hat{G}) = \text{arg}\min\limits_{W,G} \left\{ \frac{1}{2}\|Y-XW-G\|_F^2+\lambda\|W\|_{1,2}+\rho\|G\|_1 \right\}.
\end{align}
Note as pointed out in \citep{Liu2014}, there are some limitations in the least square loss function in \eqref{eq:OMR_Gross}.


To illustrate the applicability of model~\eqref{eq:CMR_Gross}, we use 3D hand pose estimation from depth images as an example. In such context, each instance is a depth image of human hand and the associated observation is a vector containing $(x, y, z)$ coordinates of all hand joints. The total length of the observation vector depends on the number of joints. It is clear that entries in the observation vector are distinct and yet \emph{intrinsically connected} in the sense that they describe 3D coordinates of different joints, but joints in the same finger are connected. Moreover, the noise levels of these entries are not necessarily the same. For example, finger tips are prone to have large error while other joints of finger have smaller error. This naturally suggests the usage of multi-task regression model with CMR loss aiming to regularize different tasks with different parameters. On the other hand, ground-truth observations are obtained based on annual annotations, which is hard to be precise due to occlusions, and is truly difficult to rule out the potential existence of gross errors from either careless or malicious user annotations.

\section{Our \emph{CMRG} Algorithm}\label{sec:Algo}
Different from ordinary multivariate linear regression problems in \eqref{eq:ReqMLR} and \eqref{eq:OMR_Gross}, the optimization problem in \eqref{eq:CMR_Gross} is more challenging, as both the loss function and regularization terms are non-smooth. For this we develop a dedicated proximal ADMM algorithm which is also inline with \citep{Boyd2011,FazelPST13,SunTY15,ChenADMM13}.  
We first adopt a variable splitting procedure as of \citep{ChenSPG12,Qin2012} to reformulate \eqref{eq:CMR_Gross} as an equivalent linearly constrained problem, as follows.
Let $d':=\sum_{g\in\mathcal{G}}|g|$, and denote as $\Upsilon\in\mathbb{R}^{d'\times p}$ the composition matrix from $W$, which is constructed by copying the rows of $W$ whenever they are shared between two overlapping groups. That is, provided $\mathcal{G}$ as the set of overlapping groups, we can constructed a new set $\mathcal{G'}$ of non-overlapping groups by means of a disjoint partition of $\{1,\cdots,d'\}$ conforming to the identity below
\[\mathcal{R}_{\mathcal{G}}(W)=\sum_{g\in\mathcal{G}}\|W_{g*}\|_F=\sum_{g'\in\mathcal{G'}}\|\Upsilon_{g'*}\|_F=\mathcal{R}_{\mathcal{G'}}(\Upsilon).\]
It is clear that $d' \geq d$. Moreover, the linear system $CW = \Upsilon$ explicitly characterizes the relations between $W$ and $\Upsilon$. Here $C\in\mathbb{R}^{d'\times d}$ is defined as: $C_{ij}=1$ if $\Upsilon_{i*} = W_{j*}$ and $C_{ij}=0$ otherwise.
Note $C$ here coresponds to a very sparse matrix. $D:=C^{\top}C$ denotes a diagonal matrix where each of its diagonal entries equals the number of repetitions of the corresponding row in $W$. In the special case of $C=I$, its corresponding $\mathcal{G}$ is composed of only non-overlapping groups.

Furthermore, \eqref{eq:CMR_Gross} can be equivalently reformulated as the following optimization problem
\begin{align}
(\hat{W},\hat{G}) = \text{arg}\min\limits_{(Z,\Upsilon),G,W} &~ \|Z\|_{2,1}+\lambda \mathcal{R}_{\mathcal{G'}}(\Upsilon)+\rho\|G\|_1 \nonumber \\
 \mbox{s.t.} &~ \mat Z \\ \Upsilon \rix + \mat I_n \\ 0 \rix G + \mat X \\ -C \rix W = \mat Y \\ 0 \rix, \label{eq:CMR_Gross_eqv}
\end{align}
when leting $Z=Y-XW-G$. It turns to be in the exact form of a 3-block convex problem as follows:
\begin{equation}\label{eq:3blockProb}
\min\limits_{\bs{w}, \bs{y}, \bs{z}}\{f(\bs{y})+g(\bs{z})-\left\langle \bs{b}, \bs{w}\right\rangle \mid \mathcal{F}^{\top}\bs{y} + \mathcal{H}^{\top}\bs{z} + \mathcal{B}^{\top}\bs{w}=\bs{c}\}.
\end{equation}
Here the notations are simplified as follows:
$\bs{c} := [Y^\top \ 0]^\top$, and $\mathcal{B} := [X^\top \ -C^\top]$.
$\mathcal{F} := I_{n+d'}$, $\mathcal{H} := [I_n \ 0]$, $g(\bs{z}):=\rho\|G\|_1$, $\bs{b}=0$.
$\bs{y}:=[Z^\top \ \Upsilon^\top]^\top$, $\bs{z}:=G$, $\bs{w}:=W$, $f(\bs{y}):=\|Z\|_{2,1}+\lambda \mathcal{R}_{\mathcal{G'}}(\Upsilon)$.

A natural algorithmic candidate to tackle the above-mentioned general 3-block convex optimization problem at \eqref{eq:3blockProb} or its more concrete realization as \eqref{eq:CMR_Gross_eqv}) in our context is the multi-block ADMM, which is a direct extension from the ADMM for addressing 2-block convex optimization problem~\citep{Boyd2011}.
It is unfortunately observed in \citep{ChenADMM13} that, although the usual 2-block ADMM converges, its direct extension to multi-block ADMM might however \emph{diverge}.
This non-convergence behavior of multi-block ADMM has attracted a number of research efforts for convergent variants. The study of~\citep{SunTY15} empirically examines the regime of existing multi-block ADMM convergent variants, and finds out that collectively they substantially under-performs than the direct multi-block ADMM extension that has no convergence guarantee.
Fortunately, very recently a proximal ADMM is developed in \citep{SunTY15} that enjoys both theoretical convergence guarantee as well as supreme empirical performance over the direct ADMM extension. This inspires us to propose a similar algorithm to be described below.

For optimization problem \eqref{eq:CMR_Gross_eqv}, the corresponding augmented Lagrangian function can be written as
\begin{align}\label{eq:CMR_Gross_AugLag}
\mathcal{L}_{\sigma}(Z,\Upsilon,G,W;\Lambda_1,\Lambda_2) := & \|Z\|_{2,1}+\lambda \mathcal{R}_{\mathcal{G'}}(\Upsilon)+\rho\|G\|_1 - \left\langle\Lambda_1,Z+G+XW-Y\right\rangle \nonumber \\
 & - \left\langle\Lambda_2,\Upsilon-CW\right\rangle + \frac{\sigma}{2}\|Z+G+XW-Y\|_F^2 \nonumber \\
 & + \frac{\sigma}{2}\|\Upsilon-CW\|_F^2,
\end{align}
where $\Lambda_1$ and $\Lambda_2$ are Lagrangian multipliers and $\sigma>0$ is the barrier parameter. Similar to \citep{SunTY15}, by applying proximal ADMM to solve the optimization problem of~\eqref{eq:CMR_Gross_eqv}, we obtain the following steps for updating variables and parameters at the $k$th iteration:
\begin{align}
& (Z^{k+1},\Upsilon^{k+1}) = \text{arg}\min_{Z, \Upsilon} \mathcal{L}_{\sigma}(Z,\Upsilon,G^k,W^k;\Lambda_1^k,\Lambda_2^k), \label{eq:PADMM_frame1}\\
& W^{k+\frac{1}{2}} = \text{arg}\min_{W} \mathcal{L}_{\sigma}(Z^{k+1},\Upsilon^{k+1},G^k,W;\Lambda_1^k,\Lambda_2^k), \label{eq:PADMM_frame2}\\
& G^{k+1} = \text{arg}\min_{G} \mathcal{L}_{\sigma}(Z^{k+1},\Upsilon^{k+1},G,W^{k+\frac{1}{2}};\Lambda_1^k,\Lambda_2^k), \label{eq:PADMM_frame3}\\
& W^{k+1} = \text{arg}\min_{W} \mathcal{L}_{\sigma}(Z^{k+1},\Upsilon^{k+1},G^{k+1},W;\Lambda_1^k,\Lambda_2^k), \label{eq:PADMM_frame4}\\
& \Lambda_1^{k+1} = \Lambda_1^{k}+\tau\sigma(Z^{k+1}+XW^{k+1}+G^{k+1}-Y), \label{eq:PADMM_frame5} \\
& \Lambda_2^{k+1} = \Lambda_2^{k}+\tau\sigma(\Upsilon^{k+1}-CW^{k+1}). \label{eq:PADMM_frame6}
\end{align}
Here we have turning parameters $\sigma>0$, $\tau>0$. Let $A:=X^\top X+D$, we choose initial values of $W$, $\Lambda_1$ and $\Lambda_2$ so that
\begin{equation}\label{Init_W_Lam}
\left\{
\begin{array}{l}
W^0:=A^{-1}(X^\top (Y-Z^0-G^0)+C^{\top}\Upsilon^0), \\
X^\top \Lambda_1^0-C^\top\Lambda_2^0=0.
\end{array}
\right.
\end{equation}

It turns out all the subproblems in \eqref{eq:PADMM_frame1}--\eqref{eq:PADMM_frame4} enjoy closed-form solutions. Subproblem \eqref{eq:PADMM_frame1} can be formulated as
\begin{align*}
Z^{k+1}=\mbox{arg}\min_{Z} \|Z\|_{2,1} + \frac{\sigma}{2} \|Z - \Delta_{z}^k / \sigma\|_F^2,
\end{align*}
where $\Delta_z^k:=Y-XW^k-G^k + \Lambda_1^k / \sigma$. Thus, the columns of $Z^{k+1}$ are given by
\begin{equation}\label{Update_Z}
Z^{k+1}_{*j} = \left(1- 1 / (\sigma\|(\Delta_z^k)_{*j}\|_2)\right)_{+}(\Delta_z^k)_{*j}, \ j=1,\cdots,p.
\end{equation}
Similarly, subproblem \eqref{eq:PADMM_frame1} can be formulated as
\begin{align*}
\Upsilon^{k+1} = \mbox{arg}\min_{\Upsilon} \sum_{g'\in \mathcal{G}'} \left\{\lambda \|\Upsilon_{g'*}\|_F + \frac{\sigma}{2} \|\Upsilon_{g'*} - (\Delta_{\Upsilon}^k)_{g'*}\|_F^2 \right\},
\end{align*}
where $\Delta_\Upsilon^k:=CW^k- \Lambda_2^k / \sigma$, and the solution $\Upsilon^{k+1}$ is given by
\begin{equation}\label{Update_Theta1}
\Upsilon^{k+1}_{g*} = \left(1- \lambda / (\sigma\|(\Delta_\Upsilon^k)_{g*}\|_F ) \right)_{+}(\Delta_\Upsilon^k)_{g*}, \ g\in\mathcal{G}'.
\end{equation}
To solve the subproblem \eqref{eq:PADMM_frame2}, we have
\begin{equation}\label{Update_W1}
W^{k+\frac{1}{2}} = A^{-1}(X^\top (Y-Z^{k+1}-G^k)+C^{\top}\Upsilon^{k+1}),
\end{equation}
where we used equality $X^T \Lambda_{1}^k - C^T \Lambda_{2}^k = 0$ which can be derived from equalities \eqref{eq:PADMM_frame5}, \eqref{eq:PADMM_frame6} and initial conditions \eqref{Init_W_Lam}. The solution $G^{k+1}$ to subproblem \eqref{eq:PADMM_frame3} is given by
\begin{align}\label{Update_G}
G^{k+1} = &~ \mbox{arg}\min_{G} \rho \|G\|_1 + \frac{\rho}{2} \|G - \Delta_G^k\|_F^2 \nonumber \\
= &~ \mbox{\textbf{sign}} (\Delta_G^k) \odot \max\{|\Delta_G^k| - \rho / \sigma, 0\},
\end{align}
where $\Delta_G^k=Y-XW^{k+\frac{1}{2}}-Z^{k+1} + \Lambda_1^k / \sigma$, $\mbox{\textbf{sign}}(\cdot)$ is the sign function, and $\odot$ denotes component-wise multiplication.
To solve subproblem \eqref{eq:PADMM_frame4}, we have
\begin{equation}\label{Update_W2}
W^{k+1} = A^{-1}(X^\top (Y-Z^{k+1}-G^{k+1})+C^{\top}\Upsilon^{k+1}).
\end{equation}

we are now ready to present our \textbf{Algorithm~\ref{PADMM}} for \textbf{c}alibrated \textbf{m}ultivariate \textbf{r}egression with \textbf{g}rossly corrupted observations (CMRG) that incorporates the above-mentioned components.
Note that when examining side-by-side with the direct 3-block ADMM extension, our proximal ADMM proposed as above possesses an additional step to evaluate $W^{k+\frac{1}{2}}$.
The additional cost of evaluating $W^{k+\frac{1}{2}}$ is trivial as that the inverse of $A$ is usually easy to compute (e.g., when Cholesky factorization of $A$ exists).

\begin{algorithm}[!t]
   \caption{\textbf{CMRG} (Calibrated Multivariate Regression with Gross Errors)}
   \label{PADMM}
\begin{algorithmic}[1]
   \STATE \textbf{Input:} $X$, $Y$, $\lambda>0$, $\rho>0$,  $\tau>0$, and $\sigma>0$.
   \STATE \textbf{Initialization:} $Z^0$, $\Upsilon^0$, $W^0$, $G^0$, $\Lambda_1^0$, $\Lambda_2^0$ such that $W^0:=A^{-1}(X^\top (Y-Z^0-G^0)+C^{\top}\Upsilon^0)$ and $X^\top \Lambda_1^0-C^\top\Lambda_2^0=0$. $k=0$.
   \REPEAT
   \STATE \textbf{Evaluate} $Z^{k+1}$ by \eqref{Update_Z}.
   \STATE \textbf{Evaluate} $\Upsilon^{k+1}$ by \eqref{Update_Theta1}.
   \STATE \textbf{Evaluate} $W^{k+\frac{1}{2}}$ by \eqref{Update_W1}.
   \STATE \textbf{Evaluate} $G^{k+1}$ by \eqref{Update_G}.
   \STATE \textbf{Evaluate} $W^{k+1}$ by \eqref{Update_W2}.
   \STATE \textbf{Evaluate} $\Lambda_1^{k+1}$ as well as $\Lambda_2^{k+1}$ by \eqref{eq:PADMM_frame5}.
   \STATE $k \leftarrow k+1$.
   \UNTIL{\textit{Convergence}}
   \STATE \textbf{Output:} $W^{k}$, $G^{k}$.
\end{algorithmic}
\end{algorithm}

For any optimal solution $(\hat{Z},\hat{\Upsilon},\hat{G},\hat{W})$ to problem \eqref{eq:CMR_Gross_eqv}, there exit optimal Lagrangian multipliers $(\hat{\Lambda}_{1}, \hat{\Lambda}_{2})$ such that
\begin{equation}\label{OPT_Cond}
\left\{
\begin{array}{l}
\hat{Z} + \hat{G} + X\hat{W} = Y, \ \hat{\Upsilon}=C\hat{W}, \ \hat{\Lambda}_{1}\in \partial \|\hat{Z}\|_{2,1}, \\
\hat{\Lambda}_{1}\in \rho \partial \|\hat{G}\|_{1}, \ \hat{\Lambda}_{2}\in \lambda \partial \mathcal{R}_{\mathcal{G}'}(\hat{\Upsilon}),
\end{array}
\right.
\end{equation}
where $\partial$ denotes subdifferential of convex functions. Performing the extra step is in fact crucial as it ensures the global convergence of the sequence generated by \textbf{Algorithm~\ref{PADMM}} to an optimal solution satisfying \eqref{OPT_Cond}. This is formally stated in \textbf{Theorem~\ref{ADMM_conv}}. 

\begin{theorem}\label{ADMM_conv}
Under the condition $\tau\in \left(0, (1+\sqrt{5})/2\right)$, the sequence $\{(Z^{k}$, $\Upsilon^{k}$, $G^{k}$, $W^{k}$, $\Lambda_{1}^{k}$, $\Lambda_{2}^{k})\}$ generated by \textbf{Algorithm~\ref{PADMM}} converges to a unique point $(\hat{Z},\hat{\Upsilon},\hat{G},\hat{W},\hat{\Lambda}_{1}, \hat{\Lambda}_{2})$ satisfying \eqref{OPT_Cond}, so that $(\hat{Z},\hat{\Upsilon},\hat{G},\hat{W})$ is an optimal solution to optimization problem \eqref{eq:CMR_Gross_eqv} and $(\hat{\Lambda}_{1}, \hat{\Lambda}_{2})$ is an optimal solution to the dual problem of \eqref{eq:CMR_Gross_eqv}.
\end{theorem}

\begin{proof}
The proof can be derived by applying Theorem 2.2(iii) in \citep{SunTY15}. It is easy to verify that the solution set of \eqref{eq:CMR_Gross_eqv} is nonempty and the constraint qualification in Assumption 2.1 of~\citep{SunTY15} holds. Therefore, to complete our proof, it is sufficient to show that $\mathcal{F}\mathcal{F}^{\top}$ and $\mathcal{H}\mathcal{H}^{\top}$ in \eqref{eq:3blockProb} are positive definite, which is obvious since $\mathcal{F}\mathcal{F}^{\top} = I_{n+d'}$ and $\mathcal{H}\mathcal{H}^{\top} = I_n$.
\end{proof}

\paragraph*{Algorithmic complexity} The complexity of \textbf{Algorithm~\ref{PADMM}} consists of two main parts, corresponding to the computation of the inverse of $A$ and the updating of variables $(Z^{k},\Upsilon^{k},G^{k},W^{k},\Lambda_{1}^{k}, \Lambda_{2}^{k})$ in each iteration. Since the coefficient matrix $A$ is the same for all $k \geq 0$, one has to compute $A^{-1}$ only once before the iteration, which costs $O(d^3+nd^2)$. Moreover, when $d\gg n$, the cost can be further reduced by applying the Sherman-Morrison-Woodbury formula \citep{Golub96} $A^{-1}=(X^{\top} X+D)^{-1}=D^{-1}-D^{-1}X^{\top}(I+X D^{-1}X^{\top})^{-1}X D^{-1}$ and computing the inverse of the $I+X D^{-1}X^{\top} \in\mathbb{R}^{n \times n}$, which costs $O(n^2d+n^3)$. Thus, the cost of computing $A^{-1}$ is $O((n+d)t^2)$ with $t=\min(n,d)$. When both $d$ and $n$ are large, it might be inapplicable to compute the Cholesky factorization of $X^{\top} X+D$ or $I+X D^{-1}X^{\top}$. In this circumstance, one can solve linear systems \eqref{Update_W1} and \eqref{Update_W2} using iterative solver such as the preconditioned conjugate gradient method \citep{Golub96}. In each iteration, the cost of computing $Z^{k+1}$ is dominated by that of computing $\Delta_{z}^k$ which is $O(dnp)$. Similarly, computing $\Upsilon^{k+1}$, $W^{k+\frac{1}{2}}$, $G^{k+1}$ and $W^{k+1}$ costs $O(d'dp)$, $O(dp(n+d'+d))$, $O(dnp)$ and $O(dp(n+d'+d))$, respectively. Overall, the complexity of \textbf{Algorithm \ref{PADMM}} is $O \big((n+d)t^2 + (d'+d+n)dpN_{ite} \big)$, where $N_{ite}$ denotes the number of iterations.

\section{Empirical Evaluations}\label{sec:exp}

The central piece of our approach is the model \eqref{eq:CMR_Gross}, or CMRG in short, which is also referred to as our approach when without confusion.
There are also three related models: the OMR model in~\eqref{eq:ReqMLR} for ordinary multivariate regression,
the CMR model in~\eqref{eq:CLR} for calibrated multivariate regression, as well as the OMRG model in~\eqref{eq:OMR_Gross} for ordinary multivariate regression with gross error,
where $\mathcal{R}_{\mathcal{G}}(W)$is used instead of $\|W\|_{1,2}$.
Our proposed \textbf{Algorithm~\ref{PADMM}} is then employed in solving our proposed model, meanwhile standard ADMM is used for solving the rest models in a similar manner.

To facilitate a better understanding of the inner-working as well as a systematic evaluation of the proposed approach, we first consider a series of experiments on simulated data, where we have full access to the ground-truths, the gross errors, and the contaminated observations. This is followed by experiments on two exemplar real-world applications: Big-five personality prediction from computational psychology, and 3D hand pose estimation from computer vision. Each of these experiments is described in details in what follows.

\subsection{Simulated Experiments}\label{subsec:SynData}

We first generate simulation datasets to systematically evaluate the finite-sample performance of our new model in controlled settings.
The synthetic data are obtained following a similar scheme to that of~\citep{Liu2014}, as follows. Each dataset has 400 examples for training, 400 examples for validation, as well as $10,000$ examples for testing. More concretely, the training examples are obtained as follows:
\begin{enumerate}
\item[(1)] Each individual row of $X$ is generated by independent sampling from a 1000-dimensional normal distribution law, $N(0,\Sigma)$, with diagonals $\Sigma_{ii}=1$, and $\Sigma_{ij}=0.5$ for all off-diagonal entries $i\neq j$.
\item[(2)] Construct the structure of group sparsity as
\begin{align*}
\mathcal{G}=\{ \{1,\cdots,10\},\{6,\cdots,15\},\cdots, \{91,\cdots,100\}, \{101\},\cdots,\{1000\}\}.
\end{align*}
The regression coefficient matrix $W^{*}\in\bbR^{1000\times 13}$ are obtained by (1) for $1 \leq i \leq 100$ and and $1\leq j\leq 13$, we have $W^{*}_{ij}=(-1)^{i}e^{-(i-1)/100}$; (2) the rest entries are set to $0$.
\item[(3)] Construct the noise matrix as  $Z=BD$. Here each entry in $B\in\bbR^{400\times 13}$ is \emph{i.i.d.} sampled from zero-mean identity variance Gaussian law, $N(0,1)$; The matrix $D\in\bbR^{13\times 13}$ is a diagonal matrix, which is obtained by
\[D_0:=\sigma_{\max} I_{13}\]
or
\[ D_1:=\sigma_{\max}\cdot \text{diag}(2^{0/4},2^{-1/4},\cdots,2^{-12/4}),\]
where $D = D_0$ implies that all regression tasks contain stochastic noises of the same magnitude while $D = D_1$ implies that regression tasks contain stochastic noises of different magnitude.
\item[(4)] Construct the gross error $G^{*}\in\bbR^{400\times 13}$. The number of nonzero entries is controlled by a ratio $\gamma$ ($0 \leq \gamma \leq 1$). The positions of these non-zero entries are randomly selected, while their magnitudes are set to $\delta\cdot \sigma_{\max}$, where $\delta > 1$ is a scaling factor, and the signs are randomly assigned.
\end{enumerate}
In the same way, we can generate validation samples and testing samples except that we do not add gross errors to the validation and testing samples.

Empirical evaluations are carried out on datasets generated with different values of $\sigma_{max}$, $\gamma$ and $\delta$ to evaluate the performance of CMRG. The regularization parameters $\lambda$ and $\rho$ are obtained by
\[(\sqrt{\ln d}+\sqrt{p})*\{2^{-5},2^{-4.5},\cdots,2^{4.5},2^{5}\} \]
and
\[\{2^{-5},2^{-4.5},\cdots,2^{4.5},2^{5}\}, \]
respectively, using 5-fold cross-validation. The optimal parameter $(\bar{\lambda},\bar{\rho})$ is set by
\[(\bar{\lambda},\bar{\rho})=\text{arg}\min_{\lambda,\rho}\|\bar{Y}-\bar{X}W^{\lambda,\rho}\|_F^2.\]
Here $W^{\lambda,\rho}$ refers to the estimation obtained from parameters $(\lambda,\rho)$, while $\bar{X}$ and $\bar{Y}$ correspond to the design and observation matrices from the validation set.

The following metrics are used in our experiments:
\begin{align*}
\mathrm{Pre. Err.} & = \|\tilde{Y}-\tilde{X}\hat{W}\|_F/\|\tilde{Y}\|_F, \\
\mathrm{Adj. Pre. Err.} & = \|(\tilde{Y}-\tilde{X}\hat{W})D^{-1}\|_F/\|\overline{Y}D^{-1}\|_F, \\
\mathrm{Est. Err.}W & = \|W^{*}-\hat{W}\|_F/\|W^{*}\|_F, \\
\mathrm{Est. Err.}G & = \|G^{*}-\hat{G}\|_F/ \max\{1, \|G^{*}\|_F\},
\end{align*}
which measures prediction error on the testing data $(\tilde{X},\tilde{Y})$, adjusted prediction error on the testing data, estimation error of $W^{*}$ and estimation error of $G^{*}$, respectively. Throughout this experiment, all shown results are average results over 100 repetitions. 

\setlength{\tabcolsep}{4pt}
\begin{table}[!t]
\caption{Prediction and estimation error (in term of mean$\pm$standard deviation) of the comparison regression models: OMR, CMR, OMRG and CMRG on synthetic data generated with $D = D_0$ and $\sigma_{max} = \sqrt{2}$.} \label{table:Example1}
\centering
\scalebox{0.75}{
\begin{tabular}{|c|c|c||c|c|c|}
\hline
\multirow{2}{*}{Algorithm} & \multicolumn{2}{c||}{Without gross error} & \multicolumn{3}{c|}{With gross error ($\sigma=0.2$ and $\delta = 5$)} \\
\cline{2-6}
& Pre.Err. & Est.Err.$W$ & Pre.Err. & Est.Err.$W$ & Est.Err.$G$ \\
\hline
OMR & \textbf{0.2196$\pm$1.0e-2} & \textbf{0.2188$\pm$1.0e-2} & 0.4400$\pm$2.0e-2 & 0.4379$\pm$2.0e-2 & -- \\
CMR & \textbf{0.2196$\pm$1.1e-2} & 0.2190$\pm$1.1e-2 & 0.4326$\pm$1.9e-2 & 0.4313$\pm$2.0e-2 & -- \\
OMRG & \textbf{0.2196$\pm$1.0e-2} & \textbf{0.2188$\pm$1.0e-2} & 0.4390$\pm$1.9e-2 & 0.4360$\pm$2.0e-2 & 0.9644$\pm$2.9e-2 \\
CMRG & \textbf{0.2196$\pm$1.1e-2} & 0.2190$\pm$1.1e-2 & \textbf{0.3544$\pm$2.1e-2} & \textbf{0.3534$\pm$2.1e-2} & \textbf{0.4814$\pm$1.3e-2}  \\
\hline
\end{tabular}
}
\end{table}

\setlength{\tabcolsep}{6pt}
\begin{table}[!t]
\caption{Prediction and estimation error (mean$\pm$standard deviation) of four regression models: OMR, CMR, OMRG and CMRG on synthetic data generated with $D = D_1$ and $\sigma_{max} = \sqrt{2}$.} \label{table:Example2}
\centering
\scalebox{0.8}{
\begin{tabular}{|c|c|c|c||c|c|c|c|}
\hline
\multirow{2}{*}{Algorithm} & \multicolumn{3}{c||}{Without gross error} & \multicolumn{4}{c|}{With gross error ($\sigma=0.2$ and $\delta = 5$)} \\
\cline{2-8}
& Pre.Err. & Adj.Pre.Err. & Est.Err.$W$ & Pre.Err. & Adj.Pre.Err. & Est.Err.$W$ & Est.Err.$G$ \\
\hline
OMR & 0.1209$\pm$5.3e-3  & 0.0866$\pm$6.1e-3 & 0.1200$\pm$5.4e-3
& 0.4109$\pm$1.7e-2 & 0.4092$\pm$2.1e-2 & 0.4083$\pm$1.7e-2 & -- \\
CMR & \textbf{0.1115$\pm$4.9e-3} & \textbf{0.0612$\pm$4.4e-3} & \textbf{0.1106$\pm$4.9e-3} & 0.4052$\pm$1.8e-2 & 0.4039$\pm$2.2e-2 & 0.4032$\pm$1.8e-2 & -- \\
OMRG & 0.1175$\pm$3.6e-3 & 0.0793$\pm$3.9e-3 & 0.1120$\pm$2.6e-3 & 0.4112$\pm$1.7e-2 & 0.4078$\pm$2.2e-2 & 0.4048$\pm$2.1e-2 & 0.9800$\pm$2.5e-2 \\
CMRG & \textbf{0.1115$\pm$4.9e-3} & \textbf{0.0612$\pm$4.4e-3} & \textbf{0.1106$\pm$4.9e-3} & \textbf{0.2021$\pm$1.4e-2} & \textbf{0.1305$\pm$1.9e-2} & \textbf{0.2015$\pm$1.3e-2} & \textbf{0.2645$\pm$9.0e-3} \\
\hline
\end{tabular}
}
\end{table}

We first study the effect of stochastic noise level in different tasks by letting $D=D_0$ and $D = D_1$, and show results of four comparison models in \tablename~\ref{table:Example1} and \tablename~\ref{table:Example2}, respectively. In \tablename~\ref{table:Example1}, since metric Adj.Pre.Err. reduces to metric Pre.Err. when $D=D_0$, we do not include the results related to Adj.Pre.Err.. From \tablename~\ref{table:Example1} and \tablename~\ref{table:Example2}, we have four observations: (1) When $D=D_0$ (that is, all regression tasks contain the same level of stochastic noise) and no gross error, all four models have the same performance. (2) When $D=D_1$ (that is, regression tasks contain different levels of stochastic noise), models adopting the $\ell_{2,1}$-norm as the loss function (i.e., CMR and CMRG) outperform the ones using least square loss (i.e., OMR and OMRG) in terms of both prediction error on testing data and estimation error of $W^{*}$. (3) In the presence of gross errors, regression models OMRG and CMRG that consider gross error perform consistently better than OMR and CMR that without such consideration. (4) It is observed that CMRG usually delivers lower prediction error as well as lower estimation error of $W$ and $G$ when comparing to OMRG.
In summary, our newly proposed model CMRG achieves the best overall performance and outperforms other models by a large margin when there are gross errors.

Next, we study the effect of $\sigma_{\max}$ while letting $D = D_1$, $\sigma=0.2$ and $\delta = 5$, and show results for $\sigma_{\max} = \sqrt{2}$, $2$, $4$ in \tablename~\ref{table:Example3}. Again, we observe that regression models with calibration perform better than their counterparts without calibration, and CMRG outperforms other models by a large margin.

\begin{table}[!t]
\caption{Effect of $\sigma_{\max}$ on the prediction and estimation error (mean$\pm$standard deviation) of four regression models: OMR, CMR, OMRG and CMRG on synthetic data generated with $D = D_1$, $\gamma=0.2$ and $\delta = 5$.} \label{table:Example3}
\centering
\scalebox{0.85}{
\begin{tabular}{|c|c|c|c|c|}
\hline
Algorithm & Pre.Err. & Adj.Pre.Err. & Est.Err.$W$ & Est.Err.$G$ \\
\hline
\multicolumn{5}{c}{$\sigma_{\max}=\sqrt{2}$}\\
\hline
OMR & 0.4109$\pm$1.7e-2 & 0.4092$\pm$2.1e-2 & 0.4083$\pm$1.7e-2 & -- \\
CMR & 0.4052$\pm$1.8e-2 & 0.4039$\pm$2.2e-2 & 0.4032$\pm$1.8e-2 & -- \\
OMRG & 0.4112$\pm$1.7e-2 & 0.4078$\pm$2.2e-2 & 0.4048$\pm$2.1e-2 & 0.9800$\pm$2.5e-2 \\
CMRG & \textbf{0.2021$\pm$1.4e-2} & \textbf{0.1305$\pm$1.9e-2} & \textbf{0.2015$\pm$1.3e-2} & \textbf{0.2645$\pm$9.0e-3} \\
\hline

\multicolumn{5}{c}{$\sigma_{\max}=2$}\\
\hline
OMR & 0.5288$\pm$1.3e-2 & 0.5237$\pm$1.7e-2 & 0.5255$\pm$1.4e-2 & -- \\
CMR & 0.5157$\pm$1.6e-2 & 0.5108$\pm$1.9e-2 & 0.5132$\pm$1.6e-2 & -- \\
OMRG & 0.6200$\pm$1.8e-2 & 0.6123$\pm$1.7e-2 & 0.6058$\pm$1.8e-2 & 0.9895$\pm$2.2e-2 \\
CMRG & \textbf{0.2596$\pm$1.1e-2} & \textbf{0.1700$\pm$1.3e-2} & \textbf{0.2585$\pm$1.2e-2} & \textbf{0.2529$\pm$1.0e-2} \\
\hline

\multicolumn{5}{c}{$\sigma_{\max}=4$}\\
\hline
OMR & 1.0882$\pm$1.6e-2 & 1.0724$\pm$2.3e-2 & 1.0812$\pm$1.4e-2 & -- \\
CMR & 0.7359$\pm$2.3e-2 & 0.7319$\pm$1.9e-2 & 0.7280$\pm$2.3e-2 & -- \\
OMRG & 1.2938$\pm$4.5e-1 & 1.2672$\pm$4.6e-1 & 1.2336$\pm$5.0e-1 & 0.8554$\pm$2.3e-1 \\
CMRG & \textbf{0.3909$\pm$2.0e-2} & \textbf{0.2717$\pm$2.1e-2} & \textbf{0.3881$\pm$2.0e-2} & \textbf{0.2183$\pm$5.7e-3} \\
\hline
\end{tabular}
}
\end{table}

\begin{figure*}[!t]
\centering
    \subfigure[][Prediction error]
    {\includegraphics[width=0.4\textwidth]{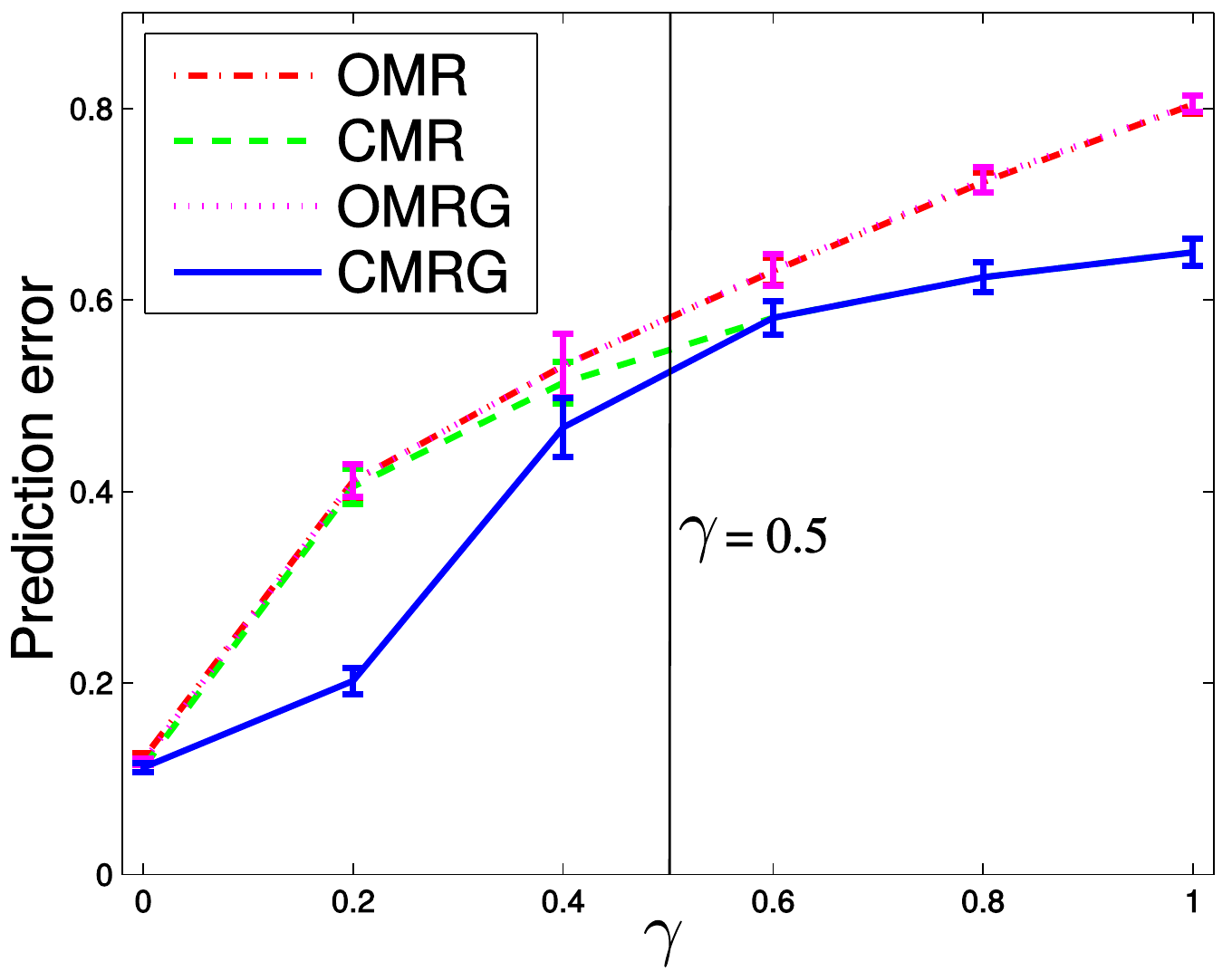}} \hspace{20pt}
    \subfigure[][Adjusted prediction error]
    {\includegraphics[width=0.4\textwidth]{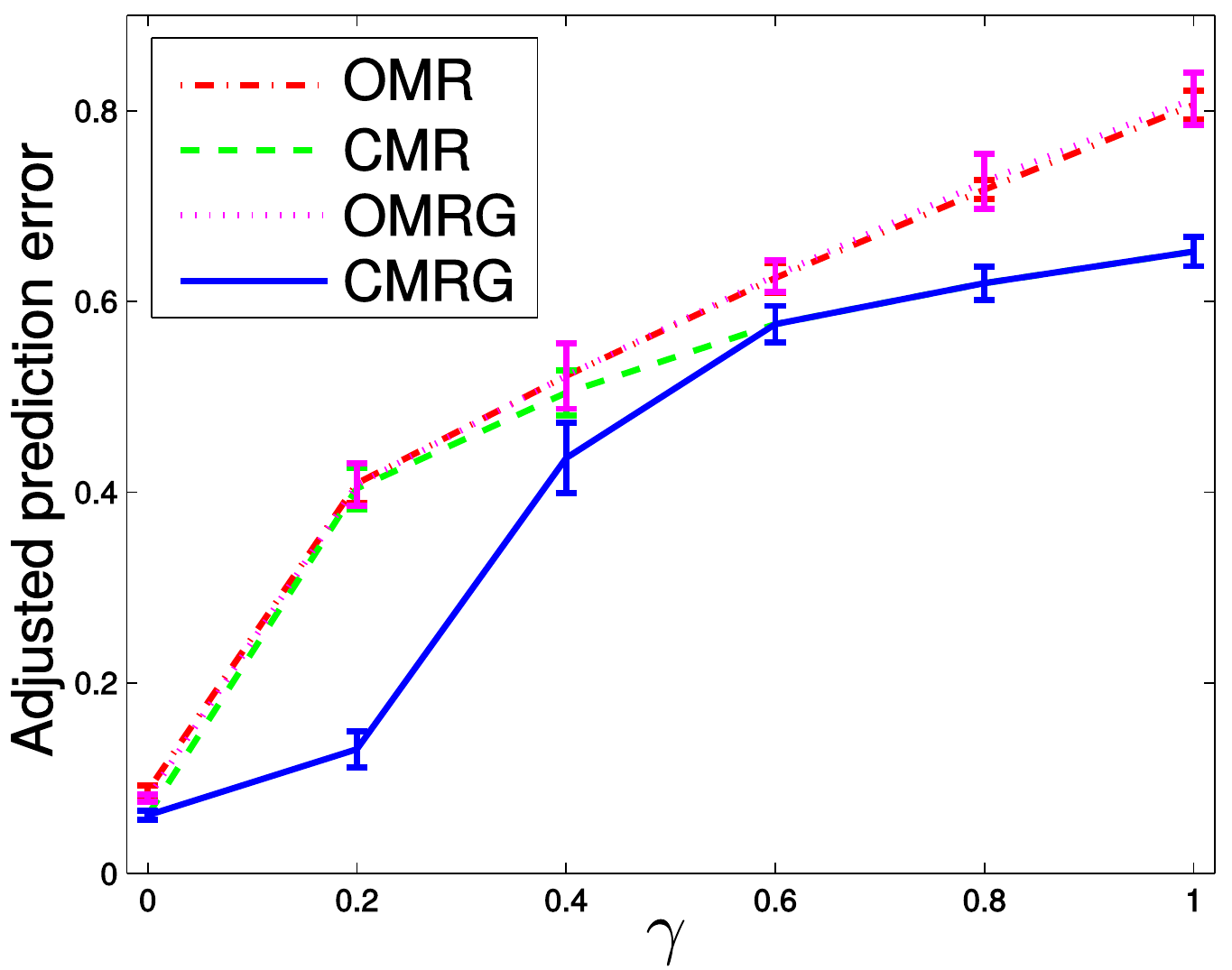}} \\
	\subfigure[][Estimation error of $W$]
	{\includegraphics[width=0.4\textwidth]{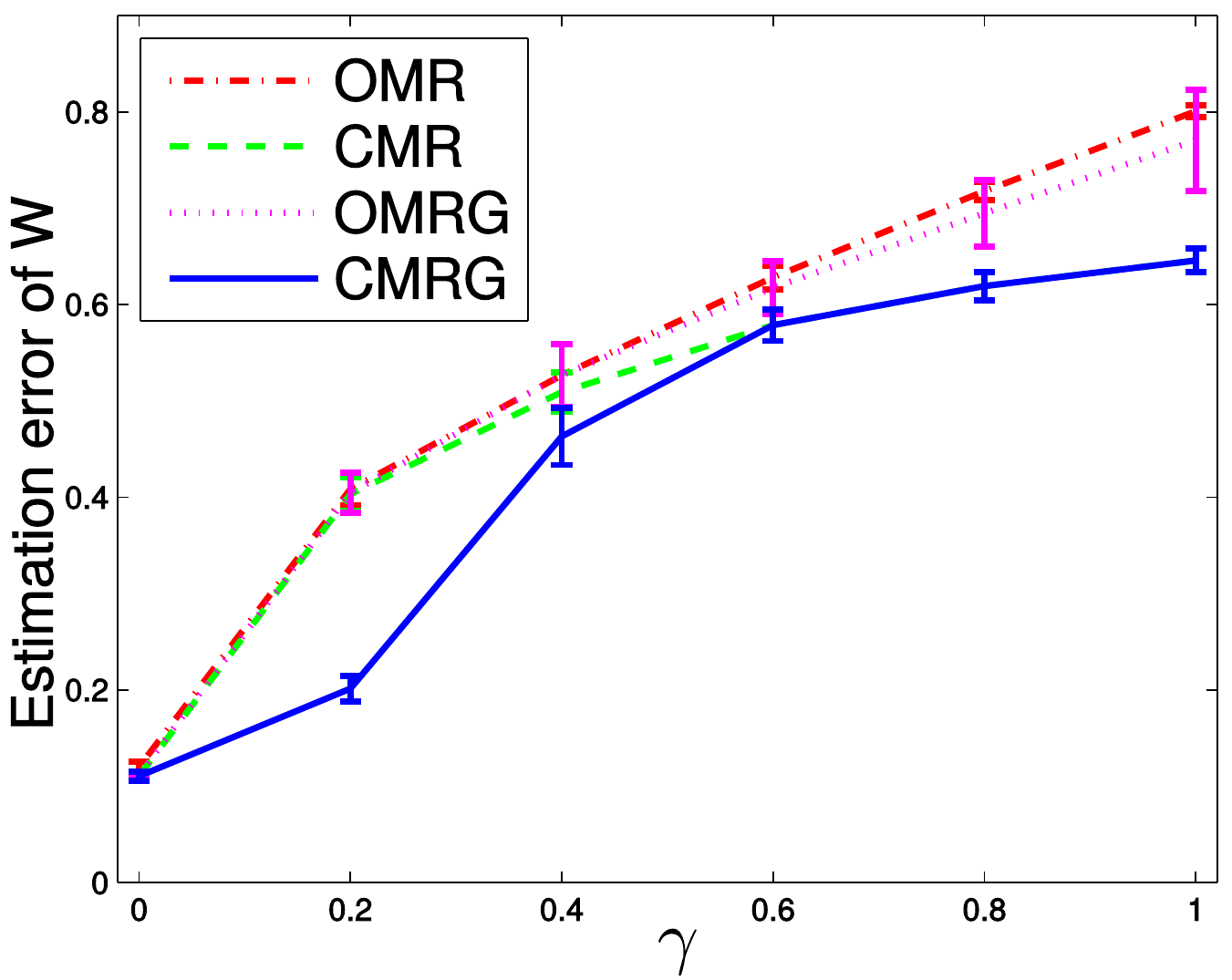}} \hspace{20pt}
	\subfigure[][Estimation error of $G$]
	{\includegraphics[width=0.4\textwidth]{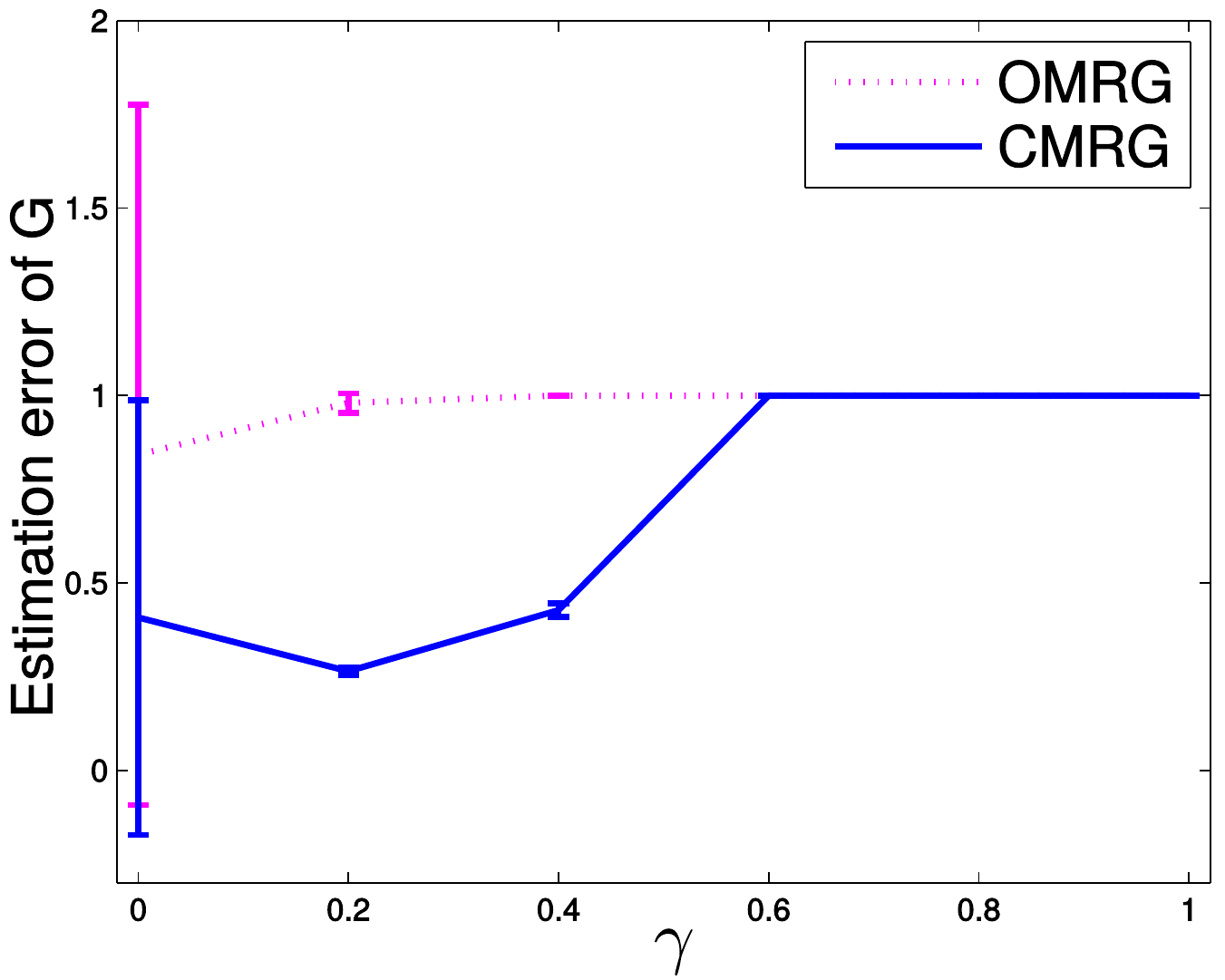}}
	\caption{Effect of the ratio ($\gamma$) of grossly corrupted training observations with $D=D_1$, $\sigma_{\max}=\sqrt{2}$, $\delta = 5$. Eeach figure shows one evaluation metric as a function of $\sigma$ with value equal to 0, 0.2, $\cdots$, 1.}\label{fig:Para_sigma}
\end{figure*}

\begin{figure*}[!t]
\centering
    \subfigure[][Prediction error]
    {\includegraphics[width=0.4\textwidth]{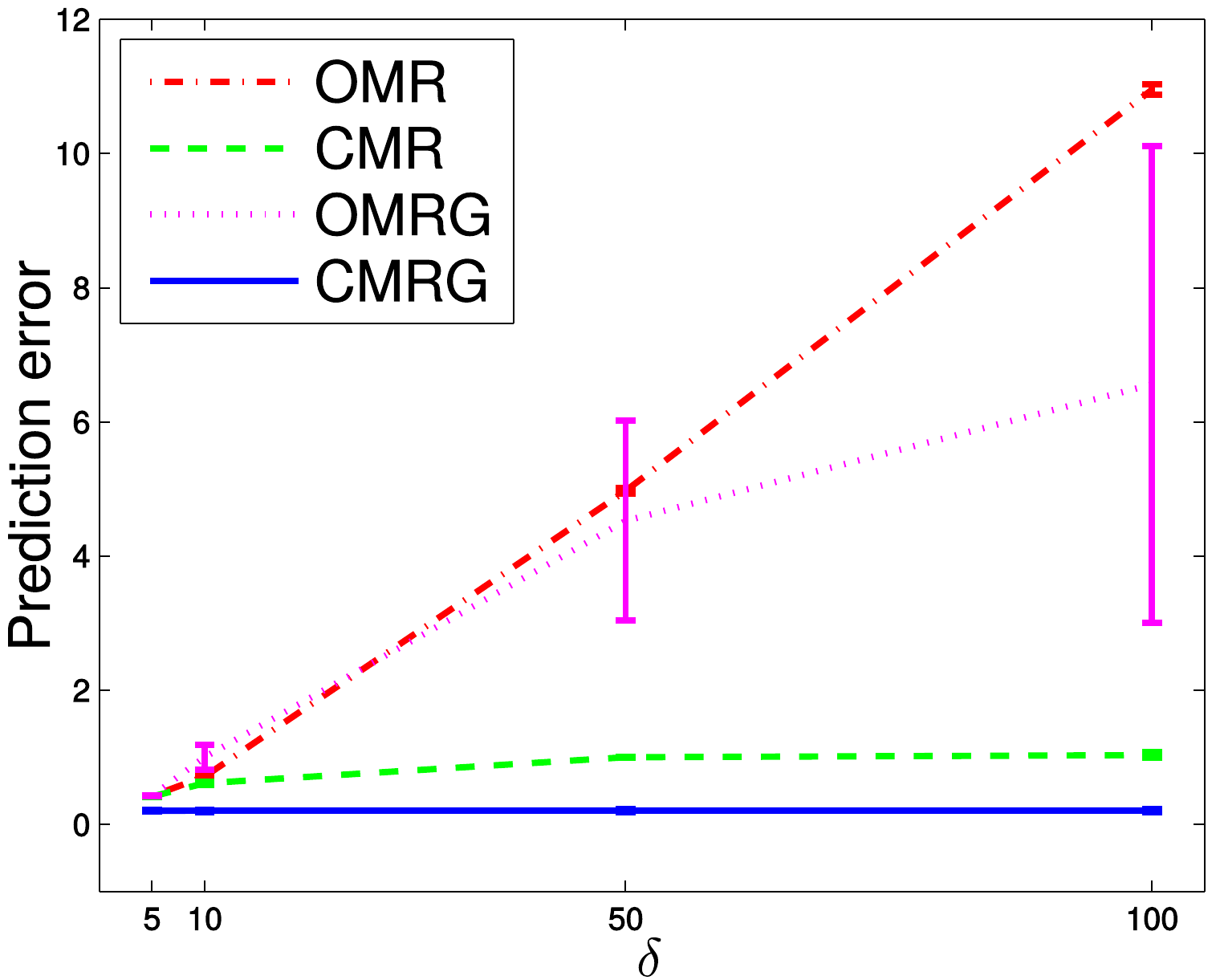}} \hspace{20pt}
    \subfigure[][Adjusted prediction error]
    {\includegraphics[width=0.4\textwidth]{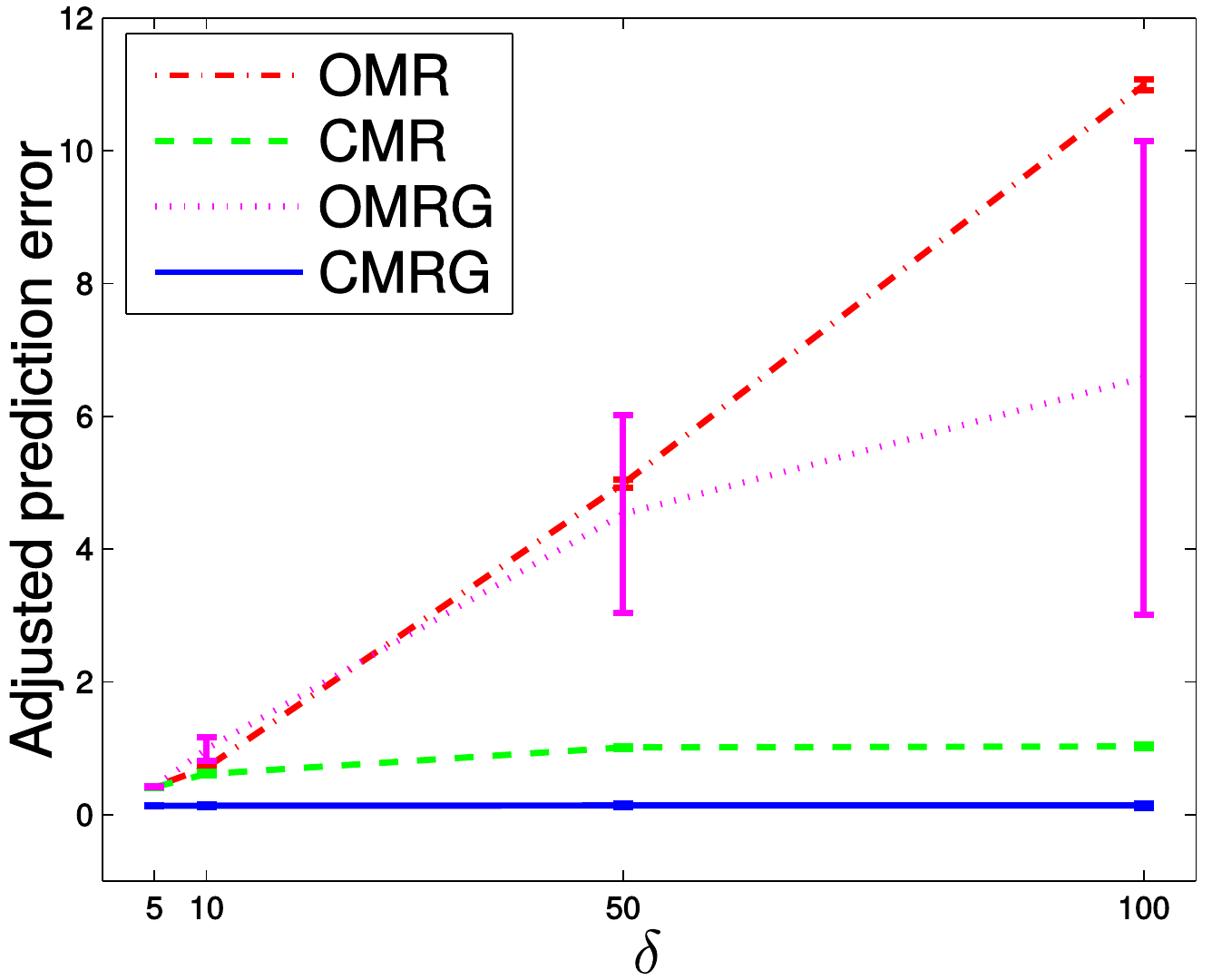}} \\
	\subfigure[][Estimation error of $W$]
	{\includegraphics[width=0.4\textwidth]{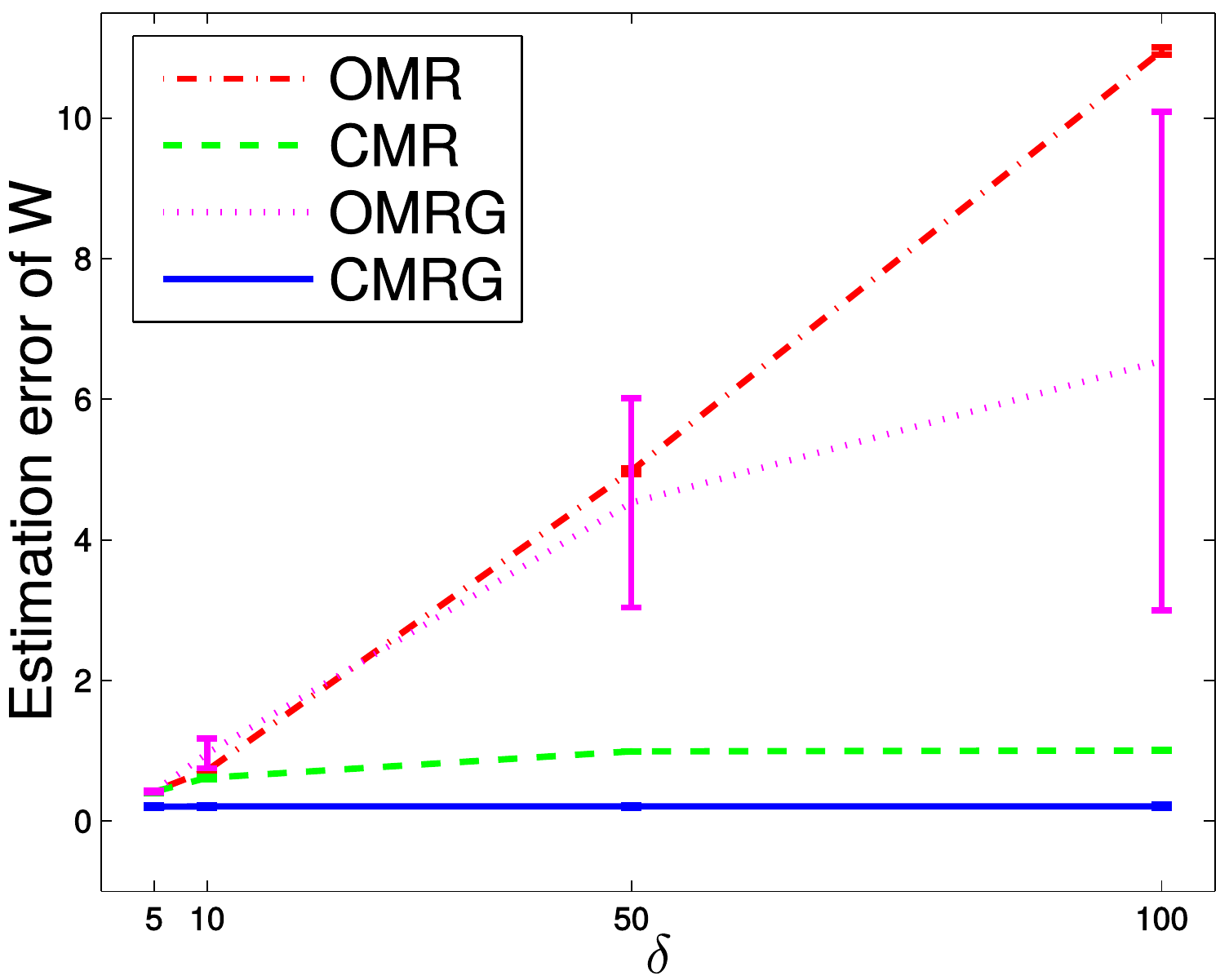}} \hspace{20pt}
	\subfigure[][Estimation error of $G$]
	{\includegraphics[width=0.4\textwidth]{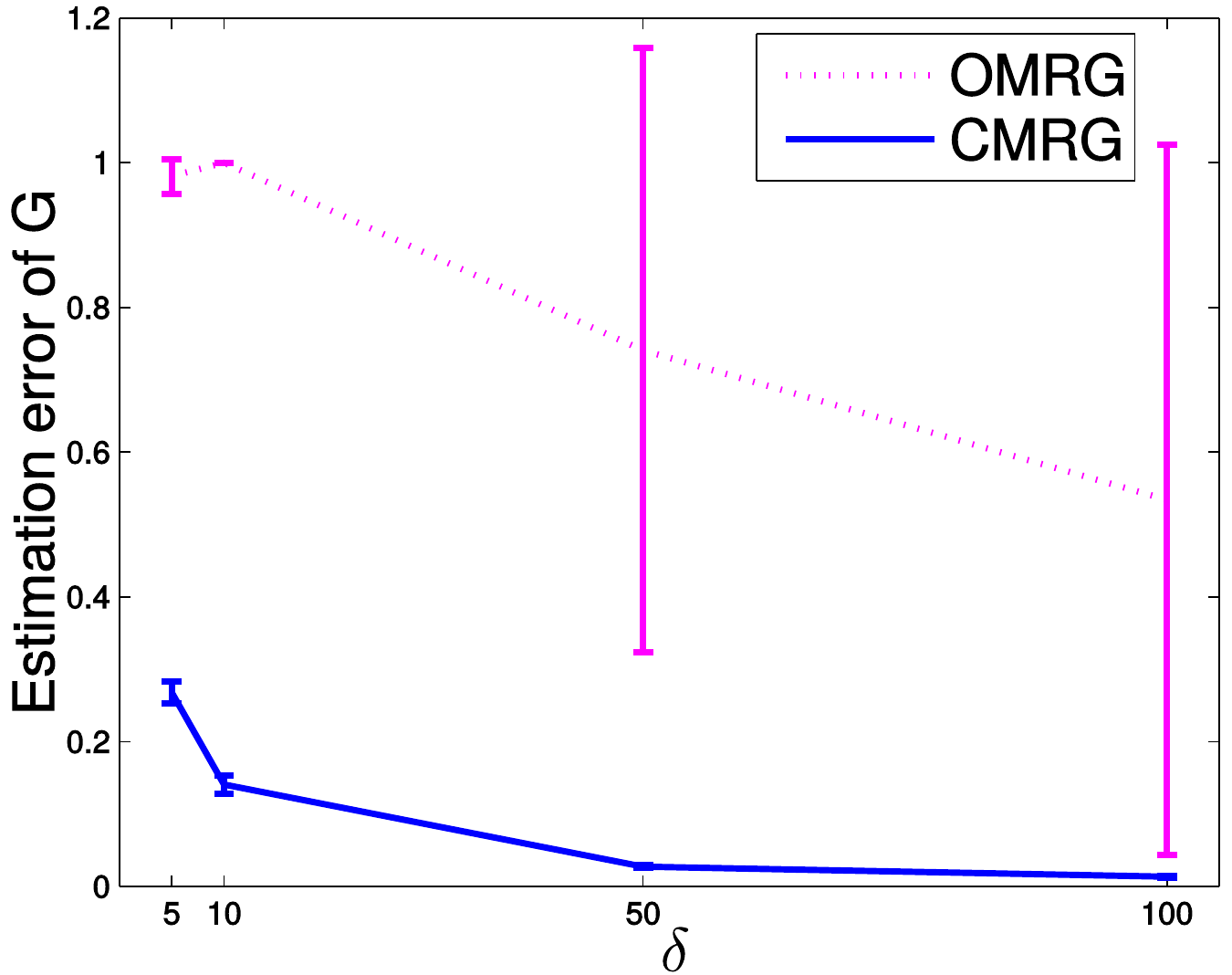}}
	\caption{Effect of the magnitude ($\delta$) of gross errors in the observations with $D=D_1$, $\sigma_{\max} = \sqrt{2}$, $\gamma = 0.2$. Eeach figure shows one evaluation metric as a function of $\delta$ with value equal to 5, 10, 50, and 100. }\label{fig:Para_delta}
\end{figure*}

We also study the effect of $\gamma$ (the ratio of gross errors in the observations) and $\delta$ (the magnitude of gross errors). \figurename~\ref{fig:Para_sigma} shows results of four regression models for $\sigma$ equal to 0, 0.2, $\cdots$, 1, and \figurename~\ref{fig:Para_delta} shows results of four regression models for $\delta$ equal to 5, 10, 50, 100. From \figurename~\ref{fig:Para_sigma}, we observe that for all four models the prediction error and estimation errors of $W$ and $G$ increase as more and more observations are grossly corrupted. Models with calibration perform better than models without calibration for all values of $\gamma$ and the advantage is more profound when $\gamma$ is large. Moreover, our newly proposed model CMRG outperforms CMR when $\gamma < 0.5$ and has similar performance as CMR when $\gamma > 0.5$ (see \figurename~\ref{fig:Para_sigma}(a)). One reason may be that CMRG fails to identify gross errors in the observations when more than half of observations are corrupted, as shown in \figurename~\ref{fig:Para_sigma}(d).
It suggests the existence of certain threshold, and our approach could recover the gross error $G^{*}$ successfully when $\gamma$ is lower than the threshold. This topic is left for future research. From \figurename~\ref{fig:Para_delta}, we see that CMRG is insensitive to the magnitude $\delta$. On the other hand, OMRG has large deviations when $\delta$ is large, which also reflects the difficulty in selecting proper regularization parameters in OMRG.

\subsection{Experiments on Personality Prediction from SNSs Behaviors}\label{subsec:BigFive}

We used a new SNSs dataset built from the microblogging site Sina Weibo (the Chinese equivalent of Twitter).
By recruiting subjects to login Weibo through our dedicated website (after filling consent forms and with legal data privacy management), these users' historical behavior data at Weibo are collected. For this Weibo dataset, 45 behavior features are constructed and are arranged into 4 groups, namely \emph{social networking}, \emph{profile}, \emph{self-presentation}, and \emph{security setting}, and in total 630 subjects are recruited for this dataset. 

This set of data is further inspected to keep only those who are active users, while reject those participants who either publish $\leq$ 512 blogs altogether, or publish zero blog during the past three months. This leads to the final dataset with only 562 subjects (instances). It is further partitioned into 450 instances for training, and 112 instances for testing. Each subject is also asked to complete a questionnaire, which is well-known BPL (Berkeley Personality Lab) Big-Five inventory consisting of forty-four inquiries. The inventory results are then epitomize into a five-dimensional personality descriptor following standard procedure. This gives rise to a vector with each element taking a value with $[1,5]$.

Four comparison methods are employed here, which include the three closely related methods (OMR, CMR, and OMRG), as well as a ridge regression or RR method, which is in fact model \eqref{eq:MLR_frame} by considering the least square loss and the regularization term of $R(W)=\|W\|_F^2$. Similar to the simulated experiments considered previously in subsection \ref{subsec:SynData}, we choose parameter $\lambda$ from $(\sqrt{\ln d}+\sqrt{p})*\{10^{-3},10^{-2.5},\cdots,10^{2}\}$, and pick-up parameter $\rho$ from the set $\{10^{-2},10^{-1.5},\cdots,10^{3}\}$. Finally, the optimal pair $(\bar{\lambda},\bar{\rho})$ is selected based on five-fold cross validation on the training set.
The relative prediction error already used during synthetic evaluations are again adopted here for performance evaluation.

To begin with, we consider evaluations w/o the presence of gross error. The left half of \tablename~\ref{WeiBo_NoGross} illustrates averaged results over 10 repetitions, where personalities \emph{Agreeableness}, \emph{Conscientiousness}, \emph{Extraversion},
\emph{Neuroticism} and \emph{Openness} in the left-most column denote the average prediction error evaluated for the corresponding personality. Further, \emph{Pre.Err.} stands for the relative prediction error averaged over the 5 output personalities.
As displayed in Table \ref{WeiBo_NoGross}, empirically OMRG and CMRG (i.e. the two models that consider gross error) performs on par with OMR and CMR (i.e. the other two methods that do not consider gross error at all), in the current dataset context where there is no gross error). Furthermore, CMRG outperforms OMR and OMRG, which is slightly taken over by RR.

\begin{table*}[!t]
\caption{Performance of five competing methods on Weibo data.} \label{WeiBo_NoGross}
\centering
\begin{tabular}{|c|ccccc|ccccc|}
\hline
\multirow{2}{*}{Personality} & \multicolumn{5}{c|}{Without gross error} & \multicolumn{5}{c|}{Corrupted data (with 10\% missing observations)} \\
 \cline{2-11}
& RR & OMR & CMR & OMRG & CMRG & RR & OMR & CMR & OMRG & CMRG \\
\hline
\textit{Agreeableness} & 0.1784 & 0.1788 & \textbf{0.1783} & 0.1788 & \textbf{0.1783} & 0.2176 & 0.2146 & 0.2136 & 0.2055 & \textbf{0.1914} \\
\textit{Conscientiousness} & \textbf{0.2128} & 0.2226 & 0.2212 & 0.2226 & 0.2212 & 0.2332 & 0.2174 & 0.2170 & 0.2160 & \textbf{0.2109} \\
\textit{Extraversion} & \textbf{0.2147} & 0.2172 & 0.2152 & 0.2172 & 0.2152 & 0.2384 & 0.2340 & 0.2379 & 0.2310 & \textbf{0.2205} \\
\textit{Neuroticism} & \textbf{0.2262} & 0.2269 & 0.2271 & 0.2269 & 0.2271 & 0.2670 & 0.2676 & 0.2622 & 0.2594 & \textbf{0.2543} \\
\textit{Openness} & \textbf{0.1717} & 0.1830 & 0.1823 & 0.1830 & 0.1823 & 0.2088 & 0.1822 & 0.1814 & 0.1720 & \textbf{0.1641} \\
\hline
Overall Pre.Err. & \textbf{0.1993} & 0.2046 & 0.2037 & 0.2046 & 0.2037 & 0.2320 & 0.2231 & 0.2221 & 0.2166 & \textbf{0.2076} \\
\hline
\end{tabular}
\end{table*}

In practice, there are situations where some entries in the multivariate output space are missing. To further investigate this type of cases, we consider a processing of our personality dataset where $ 10\%$ entries (which amounts to $250$ out of the total output entries) in the observations are randomly replaced by zero (i.e. they are deleted). Experiments are then carried out based on this corrupted dataset (for both training and testing). The right half of \tablename~\ref{WeiBo_NoGross} presents average prediction errors over ten repeats, where we clearly observe that CMRG significantly  outperforms the rest competitors. To further investigate the ability of our newly proposed model in identifying gross errors in the observations, we introduce an additional metric Rec.Rate.$G$ which quantifies the fraction of perfectly restored positive / negative entry signs in $G^*$. In other words, Rec.Rate.$G$ equals to the number of entries in $\hat{G}$ and $G^{*}$ that have the same sign divided by the number of entries in $G^{*}$. \tablename~\ref{WeiBo_Recover} presents the comparison of CMRG vs. OMRG in term of restoring those missing values. Empirically CMRG is shown to be capable of accurately identifying most of the missing observations and performs much better than OMRG.

\begin{table}[!t]
\caption{Recovery accuracy of OMRG and CMRG on Weibo data with $10\%$ missing observations. } \label{WeiBo_Recover}
\centering
\begin{tabular}{|c|c|c|}
\hline
Methods & Est.Err.$G$ & Rec.Rate.$G$ \\
\hline
CMRG & \textbf{0.7130}$\pm$0.0052& \textbf{0.9920}$\pm$0.00030  \\
OMRG & 0.8859$\pm$0.2552 & 0.8923$\pm$0.0076   \\
\hline
\end{tabular}
\end{table}

In addition, since the averaged absolute distance (AAD) is widely used as an evaluation metric in the area of personality prediction, it is also applied here to measure the deviation from gold-standard to our prediction, when the corrupted data are in use. Table \ref{WeiBo_AAD} displayed the averaged results over ten repeats, from which we see that CMRG consistently outperforms other regression models.

\begin{table}[!t]
\caption{AAD results of competing methods on Weibo data with $10\%$ missing observations.} \label{WeiBo_AAD}
\centering
\begin{tabular}{|c|c|c|c|c|c|}
\hline
Personality & OMR & CMR & OMRG & RR & CMRG \\
\hline
\textit{Agreeableness} & 0.64 & 0.64 & 0.61 & 0.65 & \textbf{0.56} \\
\textit{Conscientiousness} & 0.55 & 0.55 & 0.55 & 0.59 & \textbf{0.54} \\
\textit{Extraversion} & 0.62 & 0.63 & 0.61 & 0.62 & \textbf{0.59} \\
\textit{Neuroticism} & 0.68 & 0.66 & 0.65 & 0.68 & \textbf{0.64} \\
\textit{Openness} & 0.53 & 0.53 & 0.50 & 0.61 & \textbf{0.48} \\
\hline
Average & 0.60 & 0.60 & 0.58 & 0.63 & \textbf{0.56} \\
\hline
\end{tabular}
\end{table}

\subsection{Hand Pose Estimation from Depth Images}\label{subsec:Hand}
Vision-based hand pose estimation has plenty of applications in various areas including humanoid animation, human-computer interaction, and robotic control. The core problem here is the problem of 3D hand pose estimation~\citep{ErolEtAl:cviu07,GorFlePar:pami11}, owing mostly to the complex and dexterous nature of hand articulations. Facilitated by the emerging commodity-level depth cameras such as Kinect \footnote{\url{http://www.xbox.com/en-US/kinect/}} and Softkinect \footnote{\url{http://www.softkinetic.com}}, recent efforts on 3D hand pose estimation from depth images~\citep{YeEtAl:bookchapter13,TanEtAl:cvpr14,XuEtAl:IJCV15} have led to noticeable progress in the field. In this section, we apply our CMRG method to the problem of 3D hand pose estimation from depth images. We evaluate the performance of our method on a home-grown synthesized depth image dataset as well as the benchmark NYU Hand pose dataset~\citep{tompson14tog}, which are described separately in what follows, and compare against state-of-the-art methods in this field.

\begin{figure}[!t]
\centering
\subfigure[Hand kinematic model]{\includegraphics[width=0.35\textwidth]{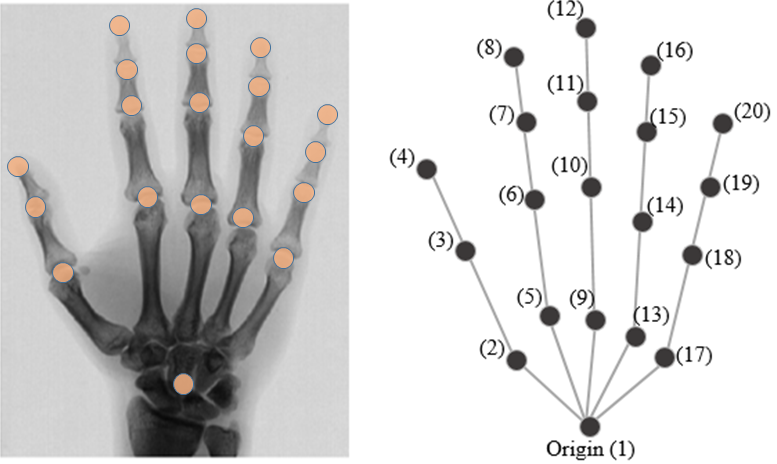}} \hspace{7pt}
\subfigure[Examples of synthesized depth image with ground-truth annotations]{\includegraphics[width=0.6\textwidth]{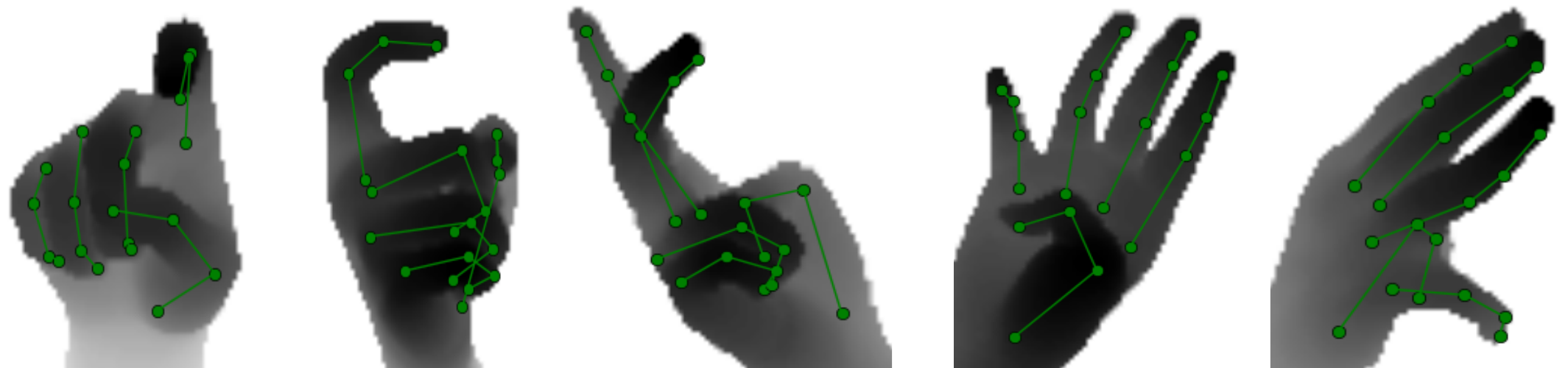}}
\caption{An illustration of the hand kinematic model and examples of hand depth image with ground-truth annotations used in our synthetic dataset for performance evaluation. For each hand image, its annotation contains 20 joints represented as a vector of length $60 = 20 \times 3$, consisting of the 3D locations of the joints following a prescribed order.}\label{fig:Hand_Kinematic}
\end{figure}

\begin{figure}[!t]
\centering
\includegraphics[width=0.6\textwidth]{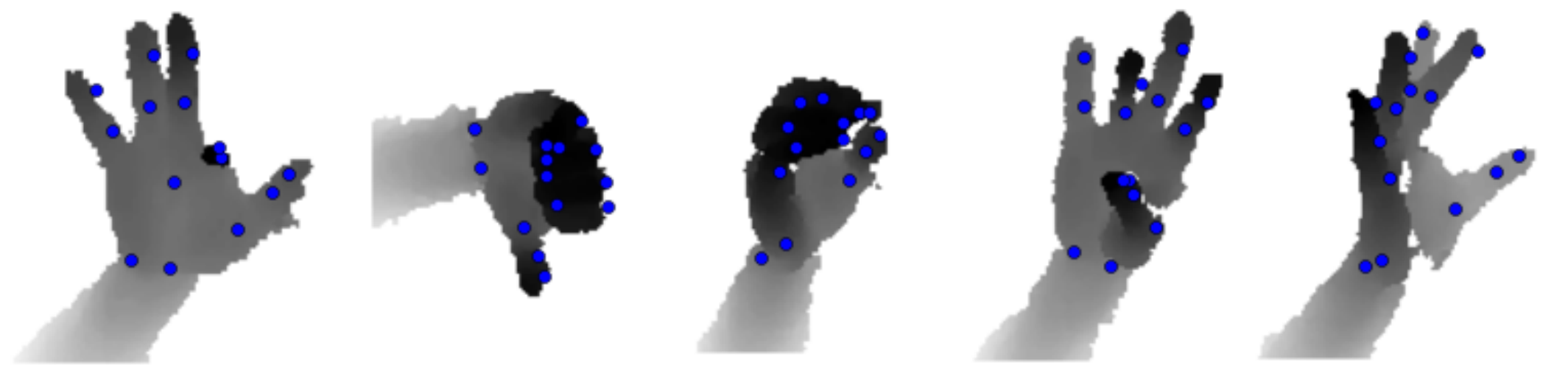}
\caption{Examples of hand depth image with ground-truth annotations from the NYU Hand pose dataset. For each hand image, its annotation contains 14 joints represented as a $42$-dimensional vector.}\label{fig:NYUHand}
\end{figure}

\begin{figure*}[!t]
\centering
\subfigure[No gross error in $Y$]{\includegraphics[width=0.3\textwidth]{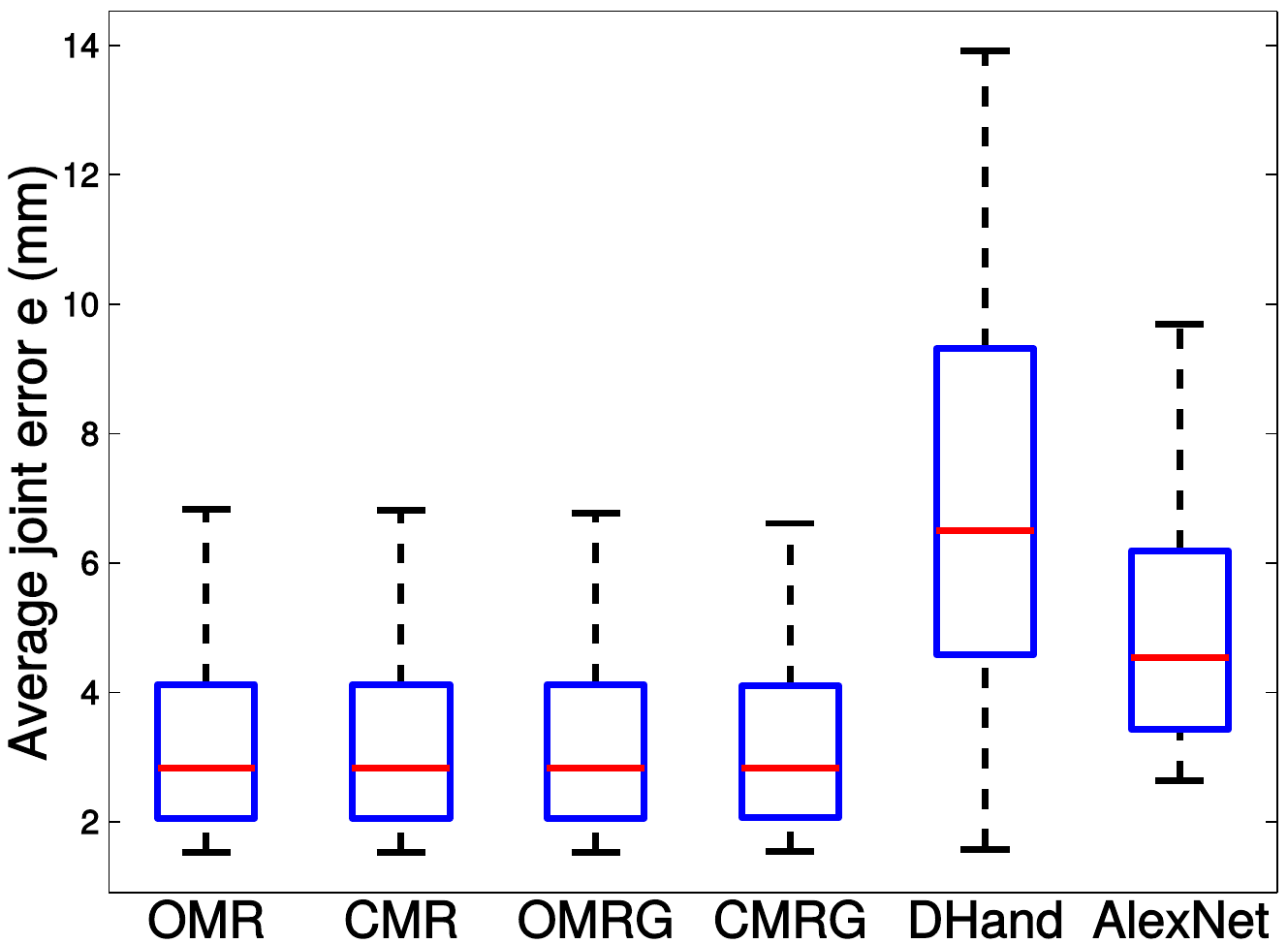}} \hspace{5pt}
\subfigure[$20\%$ entries missing in $Y$]{\includegraphics[width=0.3\textwidth]{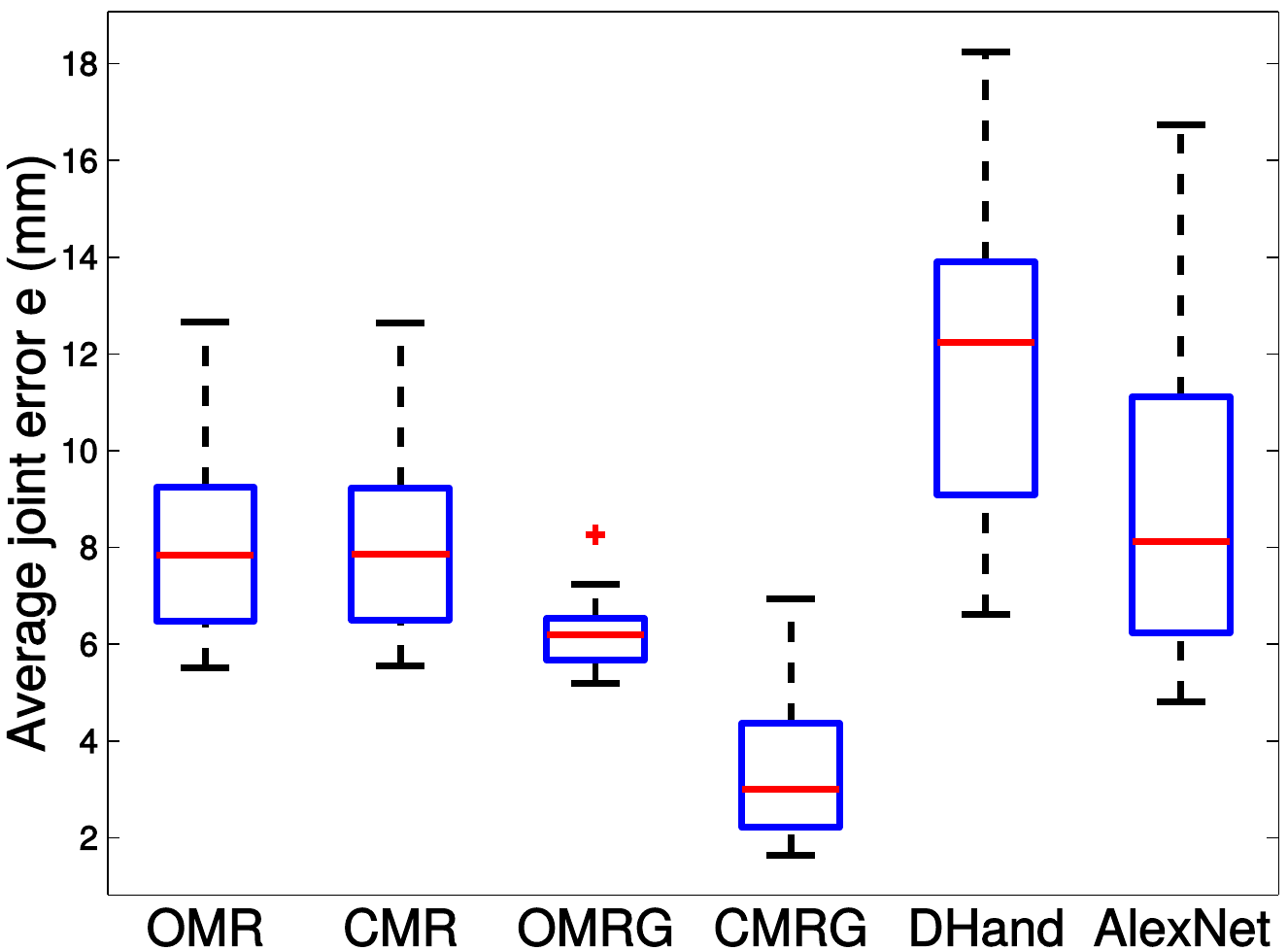}} \hspace{5pt}
\subfigure[$40\%$ entries missing in $Y$]{\includegraphics[width=0.3\textwidth]{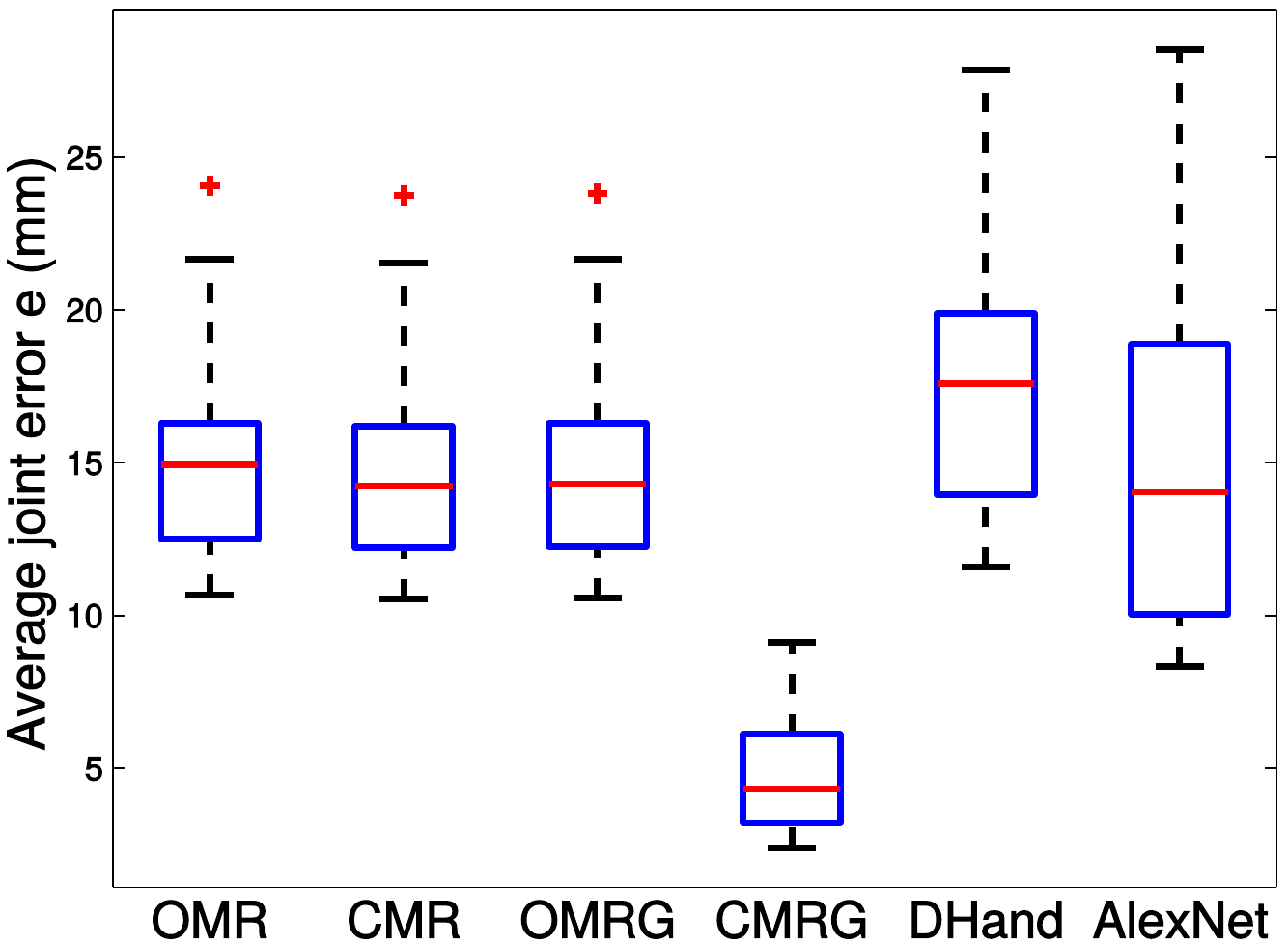}}
\caption{Average joint error $e$ in millimeter on the synthetic dataset: (a) Results using the original training data; (b) Results using the corrupted training data with $20\%$ entries in $Y$ missing; (c) Results using the corrupted training data with $40\%$ entries in $Y$ missing.}\label{fig:Hand_syn_results}
\end{figure*}

\subsubsection*{Our synthetic dataset} To conduct quantitatively analysis, we generate an in-house dataset of $110$k synthesized hand depth images, in which $80$k are used for training, $20$k are for validation, and the rest $10$k are reserved for testing. The 3D position, orientation, and hand gesture are randomly generated. The distance form a synthetic hand to virtual camera varies within the range of 650mm to 800mm. The image size obtained from the virtual depth camera is $640\times 480$, and the vertical field-of-view of the camera is 43 degree. For each depth image, an output label of hand pose is expressed in term of the set of 3D coordinates of all 20 finger joints as illustrated in~\figurename~\ref{fig:Hand_Kinematic}. We concatenate the coordinate of the joints to obtain a $60$-dimensional vector.

To apply the proposed approach, Convolutional Neural Network (CNN) features are extracted from each depth image as input $X$ in our context, and the corresponding $60$-dimensional coordinate vector corresponds to the label $Y$. The CNN features are obtained as follows: the ImageNet-pretrained AlexNet~\citep{Alex:NIPS2012} is adopted to learn a CNN model based on our aforementioned training set. Note to fulfill the input requirement of AlexNet, the depth values in each image are scaled between 0 and 255; Each image is resized properly; And each depth image is replicated three times to form a three-channel image. The MatConvNet deep learning library is adopted in this paper. The final CNN model is attained after 50 training epochs.
Now, given a new image, after applying the learned CNN model, its CNN features is obtained by simply retrieving the output from the second-to-last fully connected layer, which is a $4096$-dimensional vector.

\subsubsection*{NYU Hand pose dataset} The NYU Hand pose dataset~\citep{tompson14tog} contains 8,252 RGBD images in its test set and 72,757 in the training set~\footnote{\url{http://cims.nyu.edu/~tompson/NYU_Hand_Pose_Dataset.htm}}, from which only the depth channel images are considered in our context. Some examples of depth images are displayed in~\figurename~\ref{fig:NYUHand}.
As only 14 finger joints are annotated in the NYU dataset, here the output label $Y$ becomes a $42$-dimensional coordinate vector. Meanwhile input $X$ contains the extracted CNN features following the same protocol used in the synthetic dataset.
Now, for competing regression models OMR, CMR, OMRG and CMRG, the original 72,757 training images are randomly partitioned into two subsets: 62,757 images as training and 10,000 images as validation set to determine internal parameters $\lambda$ and $\rho$.

\begin{figure*}[!t]
\centering
\subfigure[No gross error in $Y$]{\includegraphics[width=0.3\textwidth]{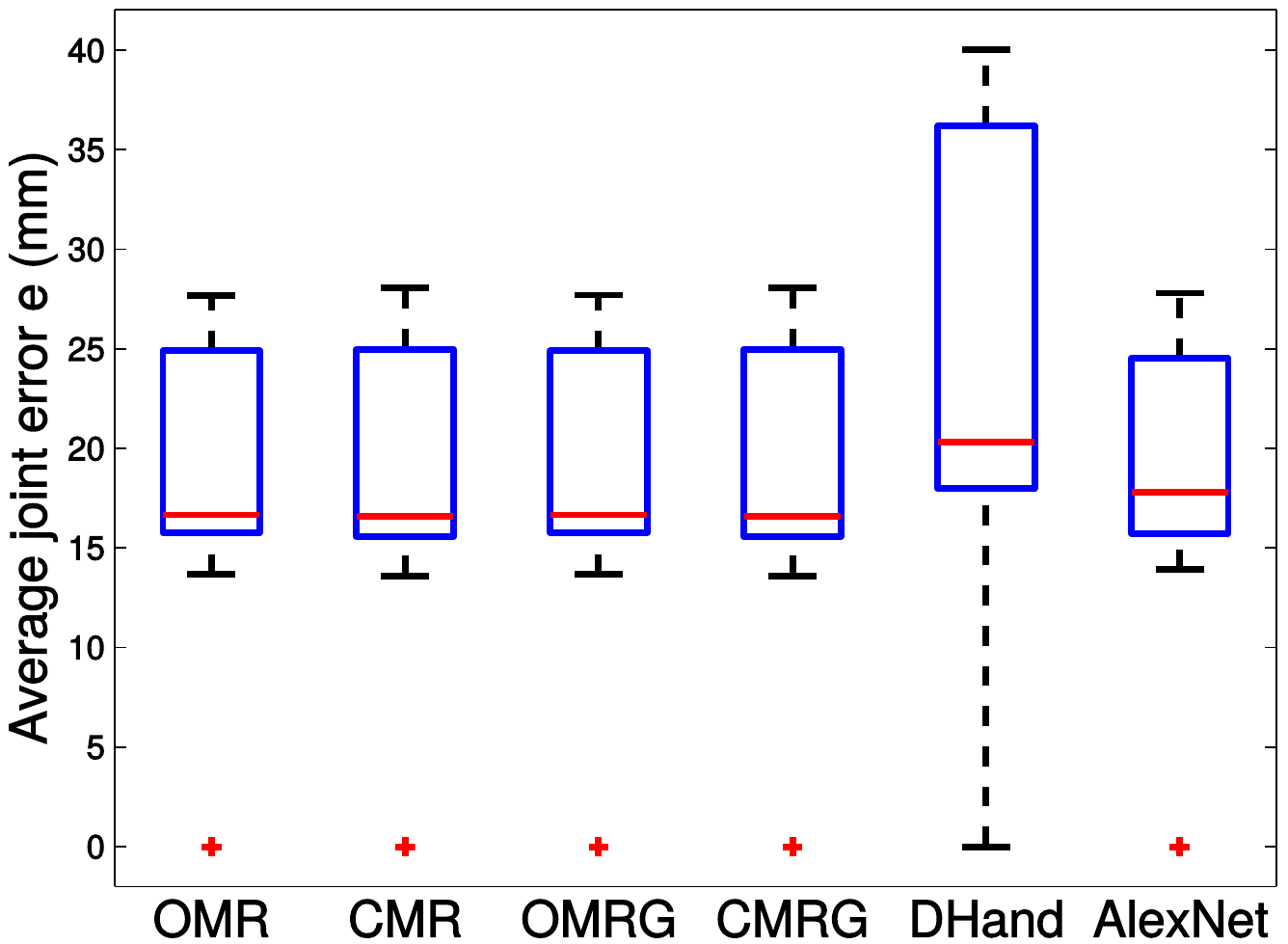}} \hspace{5pt}
\subfigure[$20\%$ entries missing in $Y$]{\includegraphics[width=0.3\textwidth]{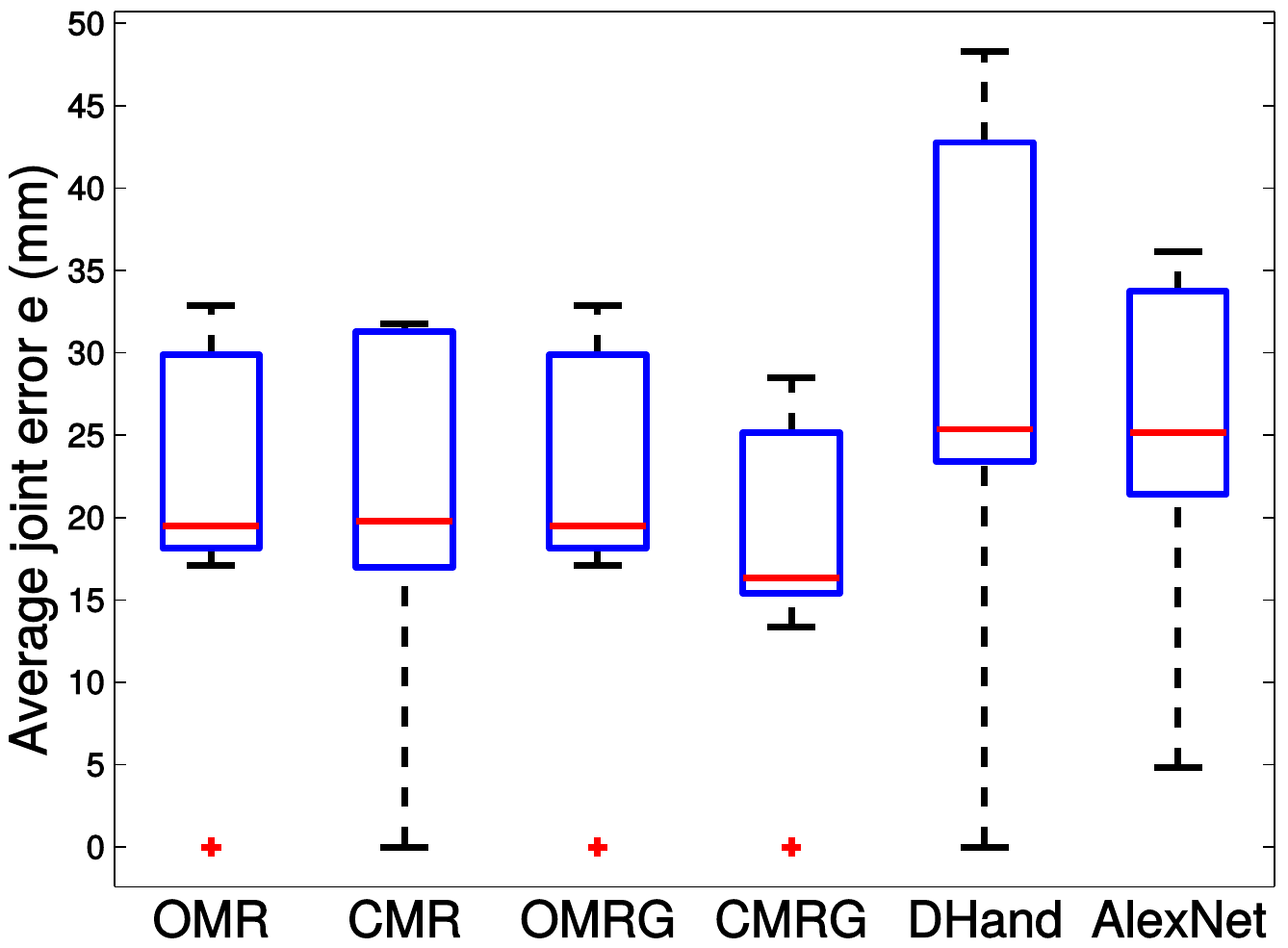}} \hspace{5pt}
\subfigure[$40\%$ entries missing in $Y$]{\includegraphics[width=0.3\textwidth]{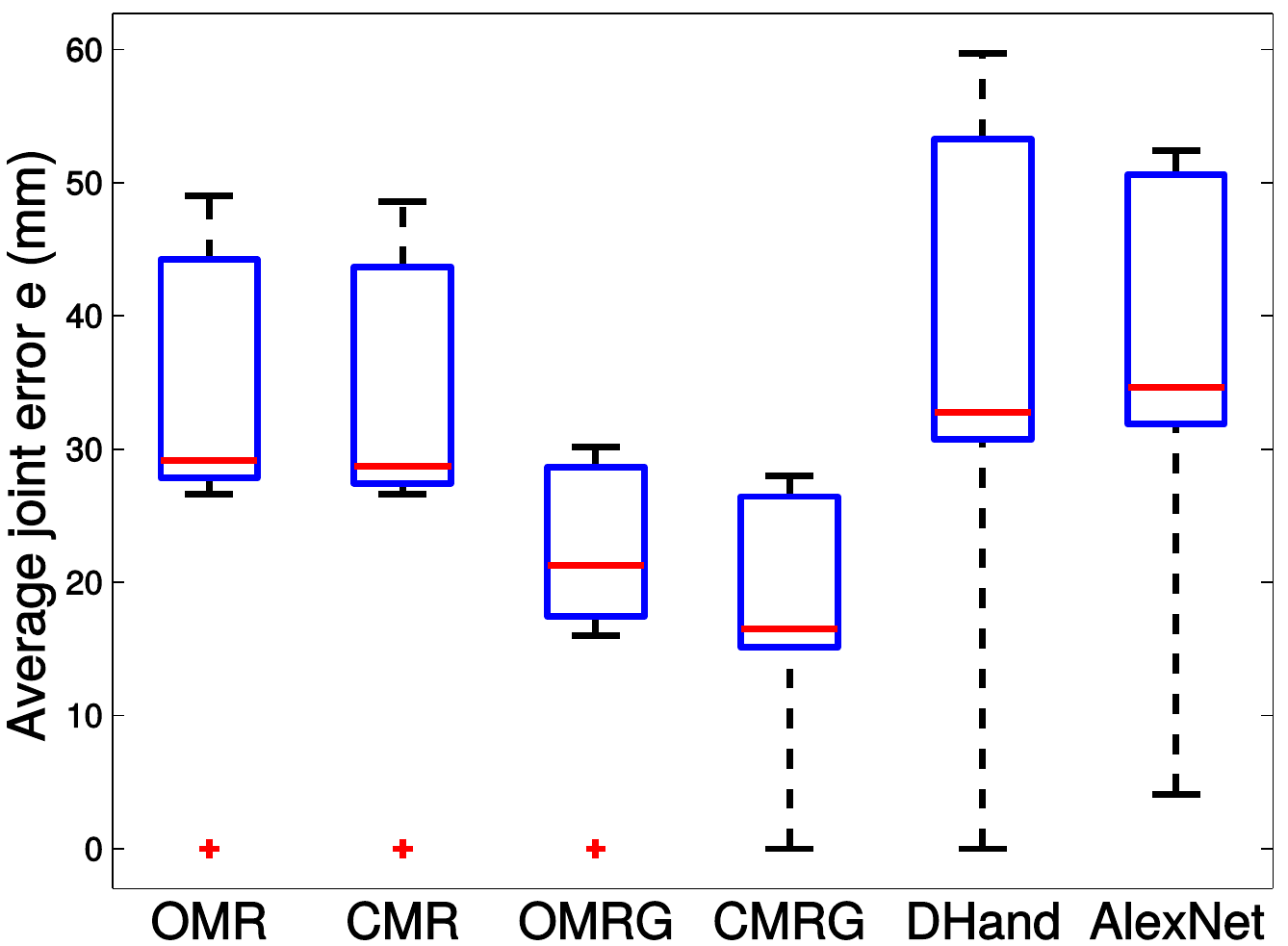}}
\caption{Average joint error $e$ in millimeter on the NYU Hand pose dataset: (a) Results using the original training data; (b) Results using the corrupted training data with $20\%$ entries in $Y$ missing; (c) Results using the corrupted training data with $40\%$ entries in $Y$ missing.}\label{fig:Hand_nyu_results}
\end{figure*}

\begin{table}[!t]
\centering
\caption{Average joint error in millimeter (mm) of the whole hand.} \label{tab:mean_joint_error}
\scalebox{0.86}{
\begin{tabular}{|c|c|c|c|c|c|c|}
\hline
 & OMR & CMR & OMRG & CMRG & DHand & AlexNet \\
\hline
\multicolumn{7}{c}{Synthetic dataset} \\
\hline
$0\%$ missing & 3.1047 & 3.1032 & 3.0995 & \textbf{3.0817} & 6.5168 & 4.9900 \\
\hline
$20\%$ missing & 8.0142 & 8.0271 & 6.2471 & \textbf{3.2753} & 11.8555 & 8.9299 \\
\hline
$40\%$ missing & 14.9866 & 14.7223 & 14.7838 & \textbf{4.6401} & 17.5957 & 14.9712 \\
\hline
\multicolumn{7}{c}{NYU Hand pose dataset} \\
\hline
$0\%$ missing & 18.3859 & 18.3527 & 18.3871 & \textbf{18.3527} & 24.8189 & 18.9210 \\
\hline
$20\%$ missing & 22.3120 & 21.9279 & 22.3106 & \textbf{18.4887} & 30.2869 & 26.0876 \\
\hline
$40\%$ missing & 33.0263 & 32.7631 & 21.3270 & \textbf{18.8763} & 38.5659 & 38.1679 \\
\hline
\end{tabular}
}
\end{table}

\subsubsection*{Evaluation metric} Following the convention of hand pose estimation literature such as~\citep{XuEtAl:IJCV15}, our performance evaluation metric is based on the \emph{joint error}, which is defined as the Euclidean distance between ground-truth and predicted 3D joint locations. Formally, denote $\hat{\mathbf{y}}_j^{i}\in \mathbb{R}^3$ and $\mathbf{y}_j^{i}\in \mathbb{R}^3$ as the ground truth and predicted joint locations for the $j$-th joint of the $i$-th testing sample. The \emph{mean joint error} of the $j$-th joint is defined as $e_{j} = \frac{1}{N_t}\sum_{i=1}^{N_t}\| \hat{\mathbf{y}}_j^{i} - \mathbf{y}_j^{i} \|$, where $N_t$ is the number of testing examples and $\| \cdot \|$ denotes the Euclidean norm in 3D space. Moreover, the mean joint error of the whole hand is simply the average of all mean joint errors, that is $e = \frac{1}{N_J}\sum_{j=1}^{N_J}e_{j}$ where $N_J = 20$ for the synthetic dataset and $N_J = 14$ for the NYU Hand pose dataset.

\subsubsection*{Experimental Set-up}
Four multivariate regression models OMR, CMR, OMRG, and CMRG are compared in the experiments. In addition, two dedicated hand pose estimation methods are considered: One is the recent work of DHand~\citep{XuEtAl:IJCV15};
For the other one, due to the recent dramatic progress of deep learning, it becomes sensible to include a CNN method based on the ImageNet-pretrained AlexNet~\citep{Alex:NIPS2012} as described earlier in the paper, where right after the 4096-dimensional fully connected layer, the standard least-square loss layer with loss term $\frac{1}{2}\|Y - XW\|_F^2$ is used for multivariate regression of joint 3D locations. Similar to the previous experiments on simulated data, for all four regression problems, the optimal parameters $\lambda$ and $\rho$ are chosen from $\{10^{-5},10^{-4.5},\cdots,10^{4.5},10^{5}\}$ based on the validation data.
Moreover, to investigate on how these methods behave when there exist missing annotations in training data, for both synthetic and NYU datasets, two additional sets of training data are obtained with $20\%$ and $40\%$ entries being randomly deleted from the ground-truth $Y$, and with the rest remains the same. Note for the deleted annotation entries, the original values are replaced by $0$.

\subsubsection*{Analysis}
Quantitative results of competing methods on the synthetic datasets are presented in \figurename~\ref{fig:Hand_syn_results}, where $y$-axis of each plot shows the average joint error $e$ in millimeter (mm). Mean joint error over the entire hand for all participating methods are also provided in~\tablename~\ref{tab:mean_joint_error}. From \figurename~\ref{fig:Hand_syn_results} and~\tablename~\ref{tab:mean_joint_error}, clearly our approach (CMRG) consistently outperforms the rest methods in the presence of gross errors, including domain-specific methods such as DHand, as well as deep learning baseline method. More specifically, when training with the original non-contaminated data in \figurename~\ref{fig:Hand_syn_results}(a), all four multivariate regression methods deliver similar performance, and CMRG is only slightly better. Interestingly all four methods perform better than DHand and AlexNet, which we attribute to the additional sparsity-induced regularizer adopted by all four models to enforce feature selection, in comparison with ALexNet where only the least-square empirical loss term is in use. Compared with DHand, these four regression models are fed with CNN features which may secure a performance boosting.
In particular, as illustrated during \figurename~\ref{fig:Hand_syn_results}(b)-(c), our approach stands out in term of being robust with increased missing $Y$ entries, meanwhile rest methods produce noticeably larger errors.

Similar trends can also be observed from the NYU dataset as presented in \figurename~\ref{fig:Hand_nyu_results} and~\tablename~\ref{tab:mean_joint_error}. It is worth mentioning that the advantage of our CMRG over comparison methods on the data is less significant when without gross error (in~\figurename~\ref{fig:Hand_nyu_results}(a)), but when there are increasing amount of missing entries in $Y$, CMRG behaves much better than the rest competitors by retaining a robust performance, as shown in~\figurename~\ref{fig:Hand_nyu_results}(b)-(c) and~\tablename~\ref{tab:mean_joint_error}. Inspired by the surprisingly good performance of our approach, combining our model with CNN would be an interesting direction for future research in the area of 3D hand pose estimation.

\section{Conclusions}\label{sec:concl}
We consider a new approach dedicating to the multivariate regression problem where some output labels are either corrupted or missing.
The gross error is explicitly addressed in our model, while it allows the adaptation of distinct regression elements or tasks according to their own noise levels. We further propose and analyze the convergence and runtime properties of the proposed proximal ADMM algorithm which is globally convergent and efficient. The model combined with the specifically designed solver enable our approach to tackle a diverse range of applications. This is practically demonstrated on two distinct applications, that is, to predict personalities based on behaviors at SNSs, as well as to estimation 3D hand pose from single depth images. Empirical experiments on synthetic and real datasets have showcased the applicability of our approach in the presence of label noises. For future work, we plan to integrate with more advanced deep learning techniques to better address more practical problems, including 3D hand pose estimation and beyond.


\bibliographystyle{spbasic}
\bibliography{CMR_Gross}

\end{document}